\documentclass[letterpaper,10pt,conference]{ieeeconf} %conf}  % Comment this line out if you need a4paper
\IEEEoverridecommandlockouts  
\usepackage{graphicx}
\usepackage{url}
\usepackage{amsfonts}
\usepackage{amsmath}

\begin{document}

\title{\LARGE \bf Learning Synthetic to Real Transfer\\
for Localization and Navigational Tasks}

\author{Pietrantoni Maxime, Chidlovskii Boris, Silander Tomi\\
Naver Labs Europe, France\\
firstname.lastname@naverlabs.com}

%\author{
%\IEEEauthorblockN{Pietrantoni Maxime}\\
%\IEEEauthorblockA{CentraleSup\'elec \\ France  \\
%maxpietrantoni@gmail.com}
%\and
%\IEEEauthorblockN{Boris Chidlovskii}\\
%\IEEEauthorblockA{Naver Labs Europe \\ France \\
%boris.chidlovskii@naverlabs.com}
%\and
%\IEEEauthorblockN{Silander Tomi}\\
%\IEEEauthorblockA{Naver Labs Europe \\ France  \\
%tomi.silander@naverlabs.com}
%}

%\dates{} %\today}
%\doi{}

\maketitle

\begin{abstract}
Autonomous navigation consists in an agent being able to navigate without human intervention or supervision, it affects both high level planning and low level control. Navigation is at the crossroad of multiple disciplines, it combines notions of computer vision, robotics and control. Fueled by progress in those aforementioned fields and combined with hardware and structural developments, autonomous navigation has known steady massive improvements throughout recent years. Modern techniques use realistic simulators fully leveraging modern computational power to generate experience and train models. Consequently, this raises the problematic of transferring the policy trained on the simulator to the real life which is affected by drastic imaging and actuation differences. Overcoming this reality gap remains to this day one of the main problematic that practitioners and researchers face in the robotic field. This internship aimed at creating, in a simulation, a navigation pipeline whose transfer to the real world could be done with as few efforts as possible. Given the limited time and the wide range of problematic to be tackled, absolute navigation performances while important was not the main objective. The emphasis was rather put on studying the sim2real gap which is one the major bottlenecks of modern robotics and autonomous navigation. It implies providing hands on experience, highlighting key problematics and pinpointing important aspects of the transfer task. The internship was divided into two consecutive parts, designing and training this navigation pipeline followed by tackling the transfer with the aim of making our physical agent autonomously navigate in a purposefully selected environment.  All the upstream decisions regarding the choice of the environment, the navigation framework and the design of the navigation pipeline were aimed at facilitating the subsequent transfer.  To design the navigation pipeline four main challenges arise; environment, localization, navigation and planning. The iGibson simulator is picked for its photo-realistic textures and physics engine. Likewise, Robot Operating System is integrated as soon as possible in the project to work within a realistic control environment. A topological approach to tackle space representation was picked over metric approaches because they generalize better to new environments and are less sensitive to change of conditions.  The navigation pipeline is decomposed as a localization module, a planning module and a local navigation module. These modules utilize three different networks, an image representation extractor, a passage detector and a local policy. The latters are trained on specifically tailored tasks with some associated datasets created for those specific tasks. Localization is the ability for the agent to localize itself against a specific space representation. It must be reliable, repeatable and robust to a wide variety of transformations. Localization is tackled as an image retrieval task using a deep neural network trained on an auxiliary task as a feature descriptor extractor. The local policy is trained with behavioral cloning from expert trajectories gathered with ROS navigation stack.  The second floor of NLE's castle was picked as the real navigation scene. It offers a diverse space layout with challenging areas both to localize against and navigate within. As an office the scene is semantically poor with high similarities between rooms. The transfer procedure first consisted in collected enough images.  The modularity of the pipeline allowed individual transfer of different models. The passage detector was simply finetuned on a few hundred real images yielding very good accuracy on the scene. Some feature adaptation was applied to learn a real image representation extractor whose output feature distribution is the same as in simulation. Local policy was left unmodified and very few changes had to be made on the script and environments. After creating a topological map of the scene from the collected dataset, the TTbot agent was able to navigate autonomously within the scene. 
\end{abstract}

%\setboolean{displaycopyright}{true}

%\maketitle

%\thispagestyle{fancy}

%\ifthenelse{\boolean{shortarticle}}{\abscontent}{}

\section{Introduction}
\subsection{Reality gap for autonomous navigation}
Autonomous navigation consists in an agent being able to navigate without human intervention or supervision, it affects both high level planning and low level control. Navigation is at the crossroad of multiple disciplines, it combines notions of computer vision, robotics and control. Fueled by progress in those aforementioned fields and combined with hardware and structural developments, autonomous navigation has known steady massive improvements throughout recent years. The most iconic example being car autonomous navigation whose dynamic ecosystem combines major technological companies, traditional car manufacturers, start ups and university labs with some advertised successes. Multiple other fields such as industry or military benefit from autonomous navigation application and contribute to its development. Overall two broad approaches can be identified when it comes to navigation. Firstly, traditional sequential pipelines that decompose the navigation task into consecutive and complementary sub-tasks; such as mapping, localization, planning, and local navigation. The models and engineering solution for each sub-tasks can be chosen independently based on the nature of the navigation as well as context and constraints. This approach being the older, it has been extensively studied and employed within multiple environments. Secondly, with the recent explosion of deep learning, learned end-to-end approaches are also being used. The latter exploits the structure of large amount of data and either use the reinforcement learning paradigm or the imitation learning paradigm associated with convolutional neural network. For example, combined with some memory representation structure to perform implicit mapping, deep reinforcement learning reaches human-level performances on standardized navigation tasks. However a major drawback is that these methods require extensive amount of training and experience. In real life providing such conditions is often intractable, either due to time limit or hazard for the agent and its environment. Therefore, a logical alternative is using a simulator alleviating both constraints and fully leveraging modern computational power. Consequently, this raises the problematic of transferring the policy trained on the simulator to the real life which is affected by drastic imaging and actuation differences. Overcoming this reality gap remains to this day one main problematic that practitioners and researchers face in the robotic field.

\subsection{Internship objectives and associated problematics}

Designing a functional autonomous navigation pipeline within a simulated environment, training it and transferring it to the real world is the main objective of this internship. This sim2real problematic will condition many choices in order to facilitate the transfer and try to make it effective with as few resources and modifications as possible. Under this orientation four main critical challenges arise; environment, localization, navigation and planning. Using Naver Labs Europe resources, the software environment must enable us to solve the engineering issues related to software, robotics and controlling the robot both in simulation and real life in a unified manner. Localization is the ability for the agent to localize itself against a specific space representation. It must be reliable, repeatable and robust to a wide variety of transformations. Navigation consists in controlling robot motors while following a policy that incorporates desired behaviors. Finally planning utilizes and combines the last two capabilities to solve high-level task and must show scalability and good generalization capabilities. The simulated and real environments differ in two ways, input variability and actuation. In a perfect simulation world, any visual, depth, 3D input is noiseless and sampled under perfect conditions and unified scene. However in real world any input is plagued by noise, changing conditions and non regularity. Actuation issues also inevitably arise because the simulation uses physics models that partially grasp the actual characteristics of the environment.

\section{Literature overview, framework definition}
This section offers a brief overview of popular solutions for both navigation and sim2real transfer therefore putting in perspective our framework's delimitation. It also details choices regarding the environment and agent.

\subsection{Sim2real transfer}
\subsubsection{Bridging the reality gap}
A common approach to bridge the reality gap is applying domain randomization \cite{tobin2017domain}. Domain randomization consists in providing enough simulated variability during training so that the model is able to generalize to real data. This method optimizes a stable policy across all tasks but is not suited to any specific task thus sacrificing overall performances. On the visual side domain adaptation techniques are employed to adapt models from the source simulation domain to the target real domain (developed in the next section \cite{patel2015visual,wilson2020survey}). On the control side, iterative learning control is traditionally used in robotic control problem \cite{abbeel2006using,cutler2014reinforcement,van2010superhuman}. It is a recursive online control method that relies on less calculation and requires less a priori knowledge about the system dynamics. It applies a simple algorithm repetitively to an unknown plant, until perfect tracking is achieved \cite{1636313}. Further work focus on improving the controller robustness by training the policy on different dynamic models.

Some imitation learning approaches or offline reinforcement learning approaches directly leverage the data to prepare the policy for the reality gap \cite{agarwal2020optimistic}. Despite the distribution mismatch, offline off policy reinforcement learning can utilize widely available interaction data to tackle real world problems. Meta learning which consists in creating models and algorithms that an quickly adapt to previously unknown tasks and environments . Cheap simulation experience would be use to train the meta policy to acquire a global behavior while real experience would be used to finetune the policy. This training is generally done within the model-agnostic meta learning (MAML) framework that utilizes policy gradient methods \cite{finn2017model} \cite{finn2018probabilistic}. The environment and policy being both stochastic estimating second derivates of the reward function to apply the policy gradient may prove challenging. It led to the creation of alternatives with control variates such as T-MAML~\cite{liu2019taming} or evolutionary strategies (ES-MAML~\cite{song2019maml}). 

\subsubsection{Domain Adaptation}
Domain adaptation is a particular case of transfer learning where the target and source tasks are the same while the domains differ \cite{daume2006domain,patel2015visual}. Homogeneous domain adaptation implies that source and target domains feature spaces are the same while heterogeneous domain adaptation implies the opposite. In supervised domain adaptation both labeled source and target data are available; in semi supervised domain adaptation only labeled source data is available while in unsupervised domain adaptation no labeled data is available. 

In homogeneous domain adaptation, the most common domain adaptation techniques aim at creating a domain invariant feature representation, called domain invariant feature learning. Under this domain invariant representation source and target domains would be aligned meaning that features follow the same distribution regardless of their domain \cite{zhao2018adversarial}. Thus any model that performs well on the source domain under this feature space could generalize easily to a target domain. The main assumption lies in the fact that this space exists and that the marginal labels distributions do not differ significantly. 

A first popular group of methods consists in minimizing the feature distributions between source and target domains. The divergence measure can be maximum mean discrepancy (MMD) \cite{gretton2007kernel,long2015learning}, correlation alignment \cite{sun2016deep}, constrastive domain discrepancy \cite{kang2019contrastive}, the Wassertein metric or a graph matching loss.
The second group of methods performs domain adaptation within an adversarial framework. A domain classifier is introduced, its goal is deciphering if the feature is generated from source or target data. Following standard GANs methods, the feature extractor and domain classifier are respectively trained to fool the domain classifier and correctly classify the features by alternating which module is trained \cite{goodfellow2014generative,zhao2018adversarial}. When performing back-propagation to update the feature extractor weights, adding a gradient reversal layer that negates the gradients from the domain classifier further improves performances by confusing more the domain classifier \cite{gan2017triangle,ganin2015unsupervised}. Some methods replace the domain classifier by a network that learns to approximate the Wassertein distance between distributions which is then minimized \cite{shinohara2016adversarial}.

An alternative to domain invariant feature learning is domain mapping which creates a mapping from one domain to another \cite{wu2019relgan,rozantsev2018residual,hoffman2018algorithms}. It is often performed at the pixel level through adversarial training with conditional GANs \cite{mirza2014conditional}.

\subsection{Navigation overview}
\vspace{4mm}

Navigation frameworks are extremely diversified and one could hardly identify a consensual approach. However, the main problematics related to autonomous navigation can be decomposed as follows: navigation tasks, space representation, memory representation and training paradigm.

\subsubsection{Navigation tasks and environments}
\vspace{2mm}
Navigation tasks that are being tackled in the field are wide and there does not seem to be some standard framework or task that would favored. We can draw two main categories of tasks. Either the goal is provided and limited exploration is necessary or goal is less specific and exploration is necessary.
Point goal navigation is the most common task, a goal is provided as a location or an image \cite{anderson2018evaluation,chaplot2020neural}. The latter can be complexified by adding moving obstacles and evolving within a dynamic environment. Language based instruction associated or not with landmarks images also create a navigation task \cite{anderson2018vision,bruce2018learning,kumar2018visual,xie2020snapnav}. In the RL or imitation paradigms reward functions/ expert demonstrations can be provided which replaces/alleviates the need of exploration.
By opposition the purpose of the navigation can simply be full exploration with in parallel the creation of a map \cite{chaplot2020learning}. Sometime the goal may be a high level semantic cue that has to be found through exploration.
Those two categories of tasks rely on different skill sets the former necessitating robust execution while the latter necessitates high level reasoning.
The inputs also widly differs between different frameworks. Roboticists traditionally favor LIDAR only based approaches while learned based approaches tend to rely on RGBD images or visual only input. Odometry may be used through visual odometry or wheel odometry, sometimes combined with other types of sensors such as accelerometers, GPS, magnetometers.

\subsubsection{Space representation}
\vspace{2mm}
Efficient space representation is crucial to the agent's capability to succeed, either spatial/metric or topological representations are utilized. Metric maps have been extensively used and can be constructed with any type of sensor RGB-D/LIDAR and agents can be easily localized against these representations \cite{elfes1987sonar,thrun2002probabilistic,durrant2006simultaneous}. This is the most standard approach. Non metric topological approaches originate from the idea that mammals rely mostly on landmark-based mechanisms for navigation. It presents a viable alternative to the shortcomings of metric maps which are the lack of scalability with regard to the size of the environment and the amount of experience as well as the lack of robustness to calibration issues, actuation noise and non optimal imaging conditions \cite{savinov2018semi,choset2001topological,chaplot2020neural}.
More recently, metric and topological maps have been combined at different scales to fully leverage their respective strengths \cite{thrun1998integrating,tomatis2001combining}. Some studies also combine these representations with some embedded semantic elements \cite{kuipers1991robot,bowman2017probabilistic}.

\subsubsection{Memory representation}
\vspace{2mm}
Depending on the context and task multiple approaches have been adopted, mostly using deep learning. Purely reactive architecture are used in for short-range movement that do not require complex navigation behaviors \cite{dosovitskiy2016learning,zhu2017target}. When the task becomes more sophisticated with intricate environments, unstructured general-purpose memory such as recurrent network are a viable solution \cite{zhu2017target,pathak2017curiosity,mirowski2016learning} even if they have trouble generalizing previously unseen environments . Self attention with transformer architecture has been used but it is structurally limited by its complexity when the trajectory becomes too long \cite{fang2019scene}. If the end to end approach is not sufficient, these architectures can be associated with specialized map-like representations or navigation-specific memory structure \cite{parisotto2017neural,gupta2017cognitive,chaplot2020neural,chen2019behavioral,savinov2018semi}. It may consists in learned on not spatial representations, metric maps or topological approaches. However learned spatial representation depends too much on metric consistency while topological approaches tend to rely heavily on the exploration settings.

\subsubsection{Training paradigm}
\vspace{2mm}
While the training method for navigation depends on the framework as well as the training environment we can list the following frequently used ones: reinforcement learning, imitation learning, self-supervised learning. Reinforcement learning relies heavily on a well designed reward function and must be able to handle sparse reward and is generally regarded as sample inefficient. Reinforcement learning also struggles to solve temporally extended tasks, although some work focus on combining planning and goal conditioned reinforcement learning \cite{nasiriany2019planning,eysenbach2019search}. When this reward function is hard to derive, imitation learning is an alternative relying on expert knowledge to distill behavior into a policy. Popular approaches in imitation learning are behavioral cloning which is a supervised learning task over state action pairs \cite{pomerleau1991efficient,ross2010efficient,torabi2018behavioral} and inverse reinforcement learning which learns a reward function that captures the expert behavior from expert and applies reinforcement learning afterward \cite{ng2000algorithms}. This function is generally learnt online although some approaches learn it offline by minimizing the wassertein distance between state-action distributions of the agent and the expert \cite{dadashi2020primal}. The distribution matching can also be approached in an adversarial manner in a more sample efficient way \cite{ho2016generative,fu2017learning}. Self-supervised learning has been successfully used but in a limited context.

\subsection{Navigation Framework}
\vspace{4mm}

Two elements constrained and motivated the choice of the framework. From a technical standpoint, providing the conditions for a convenient and efficient transfer was essential to answer to the problematic. From a purely practical standpoint, an internship lasts 6 months, the project was started from scratch and given the wide range of topics and problems involved time was of the essence. Also due to the nature of ROS/iGibson a wide majority of computations are to be performed locally and not on clusters.

\begin{itemize}
  \item We chose to focus solely on visual navigation ie. using only RGB images as input. While using lidar or depth map as input may have facilitated the elaboration of a navigation pipeline, focusing on RGB input allows to explore specific problematic relevant to computer vision and navigation. Lidar and depth maps also tend to be plagued by noise in the real world, while this could not be observed in simulation thus complicating potential transfer.
  \item The space will be represented as a topological map which is an intuitive way of interacting within a spatial environment. Topological representations show more robustness to actuation noise and change of imaging conditions than metric representation which are crucial characteristics when it comes to transfer. It also generalizes better to previously unseen environments independently of the scale. Thus we exclude any usage of metric distances, pose information from odometry or pose estimation. We assume there is a reliable angle estimation source which is the case in the simulation. It is also a reasonable hypothesis in real life since there are many methods to acquire reliable angle estimation : visual odometry, wheel odometry, CMUs, magnetic fields.
  \item Because the navigation policy is associated to a topological map it does not have to tackle high level decisions and adopt complex behaviours. The role of the local policy is indeed limited to going forward while avoiding obstacles thus acting as a local policy. This local policy will still guide the agent during fixed duration segments. A recurrent architecture is still required as it greatly facilitates obstacle avoidance. Despite its shortcomings we deem it sufficient to tackle navigation when combined to the other modules. 
  \item The navigation pipeline is learnt which implies that multiple learning tasks are individually derived for the modules in the pipeline. Being in a simulation, the complexity of creating a dataset and labelling the latter is not prohibitive and this will be adopted for multiple supervised classification/regression tasks. Regarding the local policy, any use of reinforcement learning is excluded because it is sample inefficient and reward functions are hardly obtainable in real-life considering an eventual transfer. Imitation learning is instead chosen which offers a good compromise between ease of implementation and performance.
\end{itemize}

\subsection{Simulation Environment and agent}
\vspace{4mm}

Four main simulation environments are typically used in autonomous navigation tasks, iGibson, Habitat-sim, Sapien, AI2Thor. iGibson and Habitat-sim use real-world scenes while Sapien and AI2Thor used fully simulated scenes, thus given our transfer problematic the latters are excluded. iGibson has a physics engine which allows for interactive assets, articulated objects and more realistic actuation while controlling the agent; which is not possible in Habitat. iGibson also offers fast visual rendering as well as physics simulation within photo-realistic texture advertised as facilitating potential transfer. We therefore chose to work within iGibson despite some software shortcomings regarding maintenance. iGibson comes with a full dataset of simulated environments containing 572 buildings, 1400 floors, 211 000 square meters of indoor space. However considering time and material constraints we will be working on a small subset of this dataset.

\begin{figure}[ht]
\centering
\includegraphics[width=0.8\columnwidth]{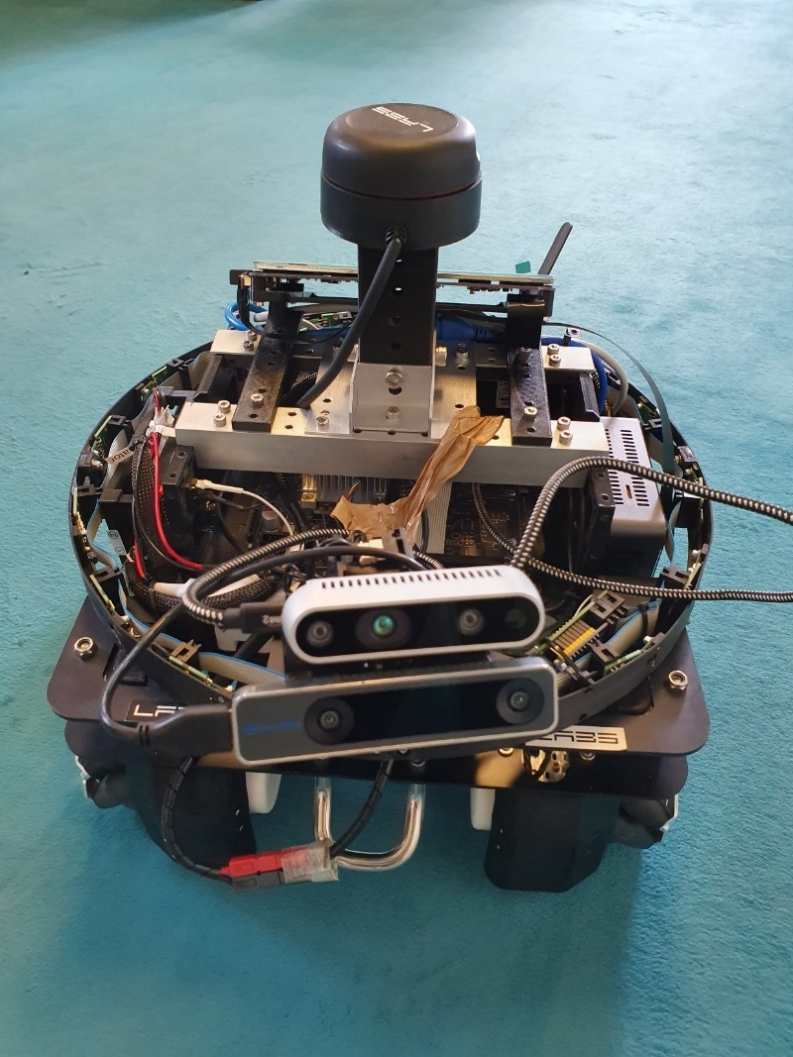}
\caption{TTbot}
\label{fig:ttbot}
\end{figure}

Common robot models are available in the base package: Fetch and Freight, Husky, TurtleBot v2, Locobot, Minitaur, JackRabbot, a generic quadrocopter, Humanoid and Ant. Throughout this work the agent used is TTbot (\ref{fig:ttbot}) which was built in-house by Naver Labs in the real world. The latter is a four wheeled-platform equipped with a Lidar, tof sensors, one standard 480*640 RGBD camera and a fisheye camera which provides a larger field of view. Two ubuntu operating systems are integrated on the platform, they use multiple homemade ROS packages to ensure full compatibility with ROS. The TTbot agent was modeled and integrated within iGibson for any simulation usage. Wheel command space is six-dimensional ($x,y,z$ linear velocity and $x,y,z$ angular velocity) providing a very wide range of motion although in practice only linear velocity $x$ and angular velocity $z$ are used.   

\begin{figure}[htbp]
\centering
\includegraphics[width=.9\linewidth]{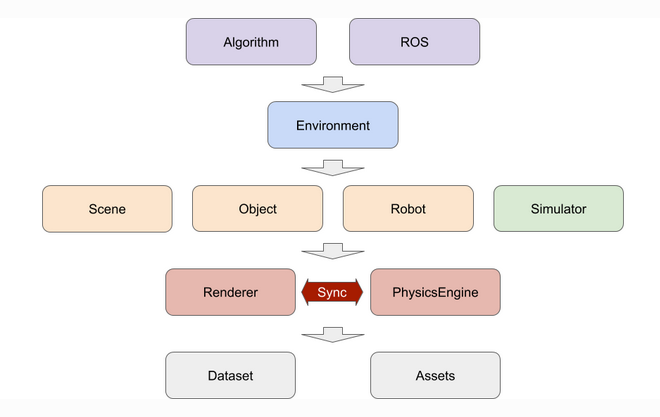}
\caption{iGibson software architecture}
\label{fig:gibson}
\end{figure}

As displayed in \ref{fig:gibson} iGibson can be decomposed in multiple hierarchical layers of abstraction, a model in a layer uses and instantiates module from layers below it. As mentioned at the bottom layer there are the dataset and assets, the dataset of ~100G contains all the 2D reconstructed real world environments while assets contain robots and objects. Built on top are the renderer and physics engine, the latter relies on PyBullet to model all the body collisions. To ensure a smooth simulation iGibson synchronizes at all time the mesh renderer and the physics engine. Next comes the simulator class which maintains an instance of the renderer and the physics engine while importing and encapsulating scene, object and robot. Finally the environment class is built on top of the simulator and follow the OpenAI gym convention while providing an API interface for applications such as ROS.

\begin{figure}[htbp]
\centering
\includegraphics[width=.95\linewidth]{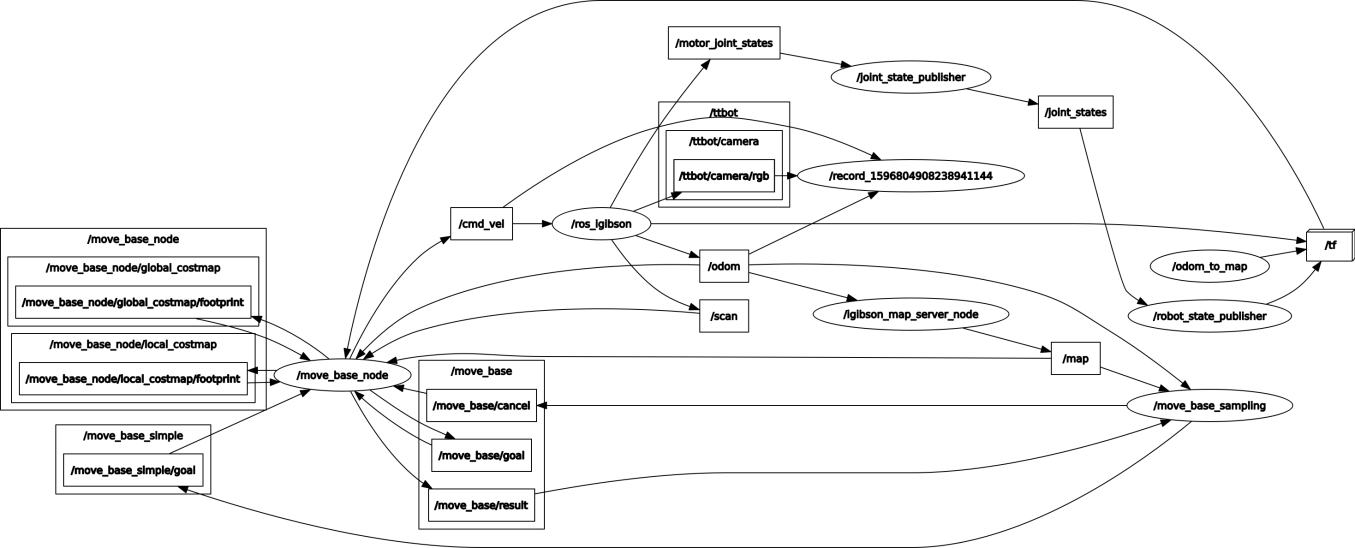}
\caption{TTbot rosgraph example}
\label{fig:rosgraph}
\end{figure}

When it comes to controlling the robot within a simulated environment two options are available. Either work within a python anaconda environment, use the iGibson python package to create the environment and control the agent or work within a catkin workspace and use ROS to control the agent while integrating the iGibson simulation within the roscore. The first option is the easiest regarding the implementation and software requirements yet by nature lacks transferability toward either another simulator or real world. Transferability affects two main points, purely technical considerations regarding integration/information flow and actuation. Throughout the internship I started using the iGibson package with python to familiarize myself with the environment. I handled visual tasks who do not differ if we use iGibson on a conda environment or through ROS. To comply to the transfer problematic mentioned earlier we quickly switched to ROS which also allowed to tackle more complex tasks. In practice within a roscore session, the iGibson environment is instantiated as a node while the python script handling the navigation is instantiated as another node, the rest is a standard ROS structure with the adapted nodes and topics as shown in Figure \ref{fig:rosgraph}. Simulation node and navigation script node do not need to interact directly except for teleportation request in which case a special code is needed. In simulation, ROS time referential is set by the iGibson simulator. In iGibson, the physics timestep as well as the action timestep are set to 240Hz. When computation or any other reason make it impossible for the simulator to keep up with this high functioning frequency, time is arbitrarily extended to match 240Hz. This implies that simulation time is much slower that real time, one simulation minute often accounts for multiple real minutes.

\section{Navigation pipeline}
Within the framework we just presented, this section introduces the whole navigation pipeline. The multiple technical challenges and navigation problematics faced and what was implemented and designed to solve them. Intermediate results will be provided for each individual module.
\subsection{Structure and associated problematic}
\vspace{4mm}

 Following standard navigation pipelines, the task is decomposed into three consecutive and complementary modules: localization, mapping and local navigation. Firstly, from the goal image provided the agent localizes the associated global goal node. Then the following navigation loop is repeated until failure or success. 
 The agent first localizes itself against the scene representation. Localization within the topological map is discrete and consists in finding the closest node to the agent's current position. Once the closest node is found, the mapping module uses the topological map to find the best path to reach the goal node. It outputs a local goal node and the absolute angle to reach it. Finally the agent orientates itself toward that angle and navigates for a fixed period of time while avoiding obstacles.    

\begin{figure}[htbp]
\centering
\includegraphics[width=1\linewidth]{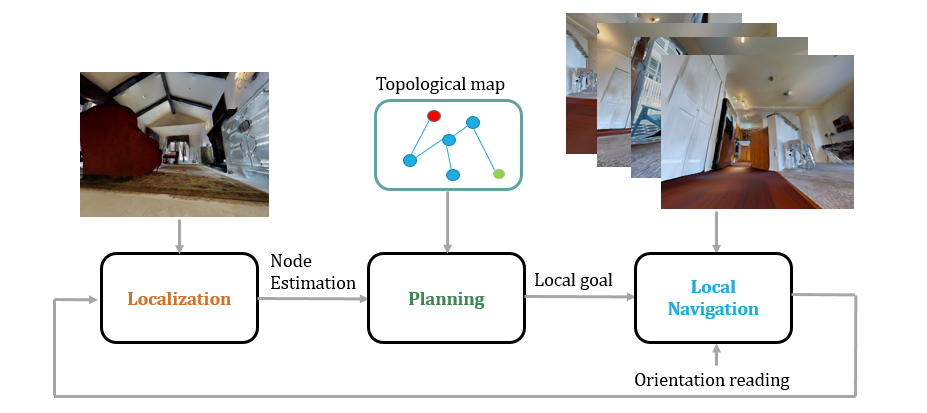}
\caption{Navigation pipeline}
\label{fig:pipe}
\end{figure}

\begin{itemize}
\item \textbf{Localization:} In an image retrieval fashion, given a query image, the agent localizes itself by finding within the database the image whose embedding share the most similarity with the query's embedding. The database contains the images collected from the topological map nodes.
\item \textbf{Planning:} Given the topological map, a goal node and a current position node; planning consists in applying Dijkstra on the graph to find the shortest path. However the edges are not associated to metric distances but to negative log-likelihood of passage which yields the most navigable path. Public python package is used for Dijkstra's shortest path algorithm.
\item \textbf{Local Navigation:} Firstly, the local navigation module uses a PID to orientate the agent toward the angle given by the planning module. Public python package is also used to implement the PID. This angle indicates the direction to follow to reach the next node, a reliable angle estimation source is available both in simulator and in real-life. Secondly the agent check that the path in front of him is navigable, if not it incrementally increases and decreases its orientation until a navigable path is found. Finally the local policy springs into action during 60s with twelve independent 5s navigation sequences.
\end{itemize}

Successful and optimal navigation requires that all modules work in conjunction and concordance without major errors. Localization must be reliable and robust to any change of imaging conditions as well as perspective change, to that end the learned image representation must be picked carefully. Failure to localize does not imply collision but will delay the agent and possibly prevent it from ever reaching the goal with consistency. The topological map space representation's must structurally allow for efficient navigation and provide all the necessary information to the other modules while respecting the aforementioned framework. Finally the local navigation module must first and foremost cover ground while avoiding collision from one time step visual input. However while avoiding collision is obvious, the local policy navigation must minimize the amount of conflict with the planning module which is a hard balance to attain. It is not meant to make planning level decision but simply act to avoid obstacle while following direction it has been given.

\vspace{5mm}
The following subsections introduce these modules, their design and training with associated motivations and some intermediate individual performance evaluation.

\subsection{Localization and image representation}
\vspace{4mm}

Efficient localization within the scene is crucial to ensure a smooth navigation. While an error in localization may not lead to immediate failure, it causes delay and increases the likelihood of collision. Moreover constant failure to localize will prevent the agent to plan and reach its goal. To prevent such scenarios we formulate the localization problem as an image retrieval task. A query image's location is estimated using the locations of the most visually similar images from a database. In image retrieval, each image is traditionally represented with locally invariant features aggregated into a single vector either as bag-of-words, VLAD or a fisher vector, indexed after reducing its dimensions.

The first challenge is deriving an image representation that would be as invariant as possible to geometric changes, robust to partial occlusion as well as robust to illumination changes. Those criterions are critical to tackle localization given that our database is relatively sparse under our topological space representation. We adopt a learning approach to leverage the power of convolutional neural network when it comes to image representation, using them as black-box descriptor extractors. The learning task used to train the network is room classification, as often the task is not directly related to image retrieval. This auxiliary task is meant to bring some distinctiveness relevant to the iGibson scenes within the features.

\subsubsection{Dataset creation}
\vspace{2mm}

Given the relative accessibility of labels and for the sake of simplicity a supervised learning approach is chosen. Given an input RGB image 480*640 the network must classify the room index the image was taken from. The project being led locally, a single gpu is available to perform the calculations, therefore we restrict the dataset to 10 scenes : Aloha, Arbutus, Beach, Elton, Foyil, Frierson, Gasburg, Natural, Rosser, Wyatt which accounts to 92 total rooms. 2 of these scenes have two floors while the rest of the scenes have 1 floor. These 92 rooms are all assigned a unique label number in an arbitrary order.

The first task is to properly segment the scenes and identify the coordinates associated to each room. The iGibson package's metadata gives the total number of rooms per scene but does not indicate where are these rooms located. Defining a room within a scene is not explicit at all depending on how the latter is defined. It may be defined by its function to the inhabitant or as a large enough area separated from another large enough area by a corridor/door. In any way these definitions are not encompassing enough to properly segment the scene into rooms, furthermore manually labelling each position would take too much time. To automatically label each position we choose to cluster all the positions within a scene using their spatial coordinates. Hence for a single scene, a arbitrarily high number of unique (x,y) coordinate pairs are sampled and used to train the clustering model. One clustering model per scene is necessary, then when creating the actual dataset the latters are used to assign the labels based on the coordinate of each position. Three clustering methodologies are tested, k-means, Gaussian mixture model and spectral clustering.
 
\begin{figure}[htbp]
\centering
\includegraphics[width=.7\linewidth]{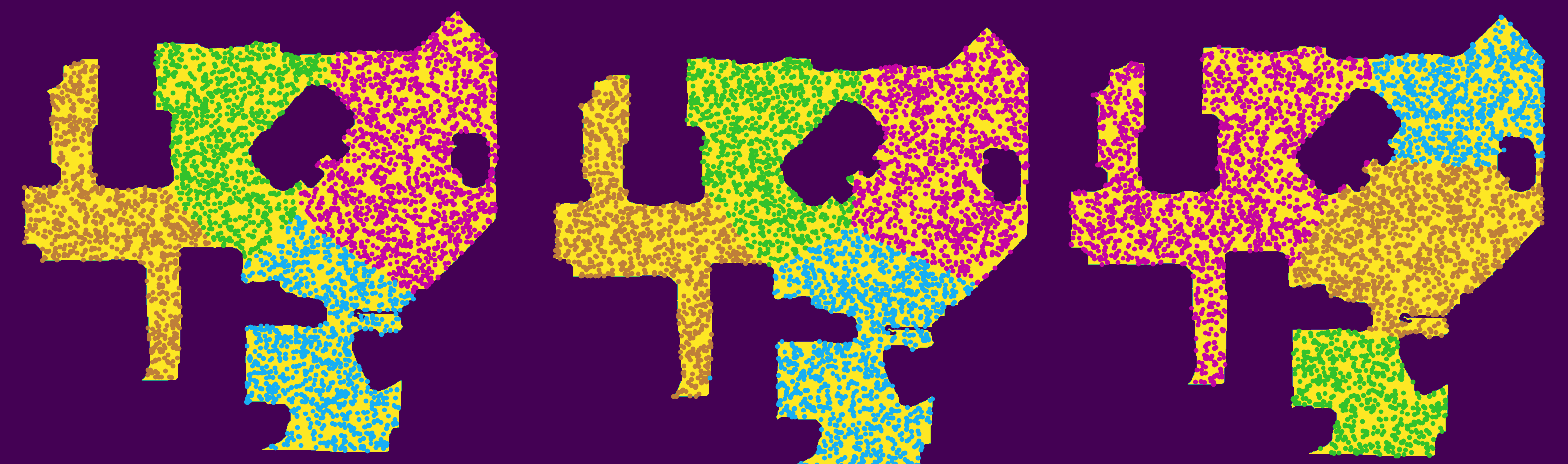}
\includegraphics[width=.7\linewidth]{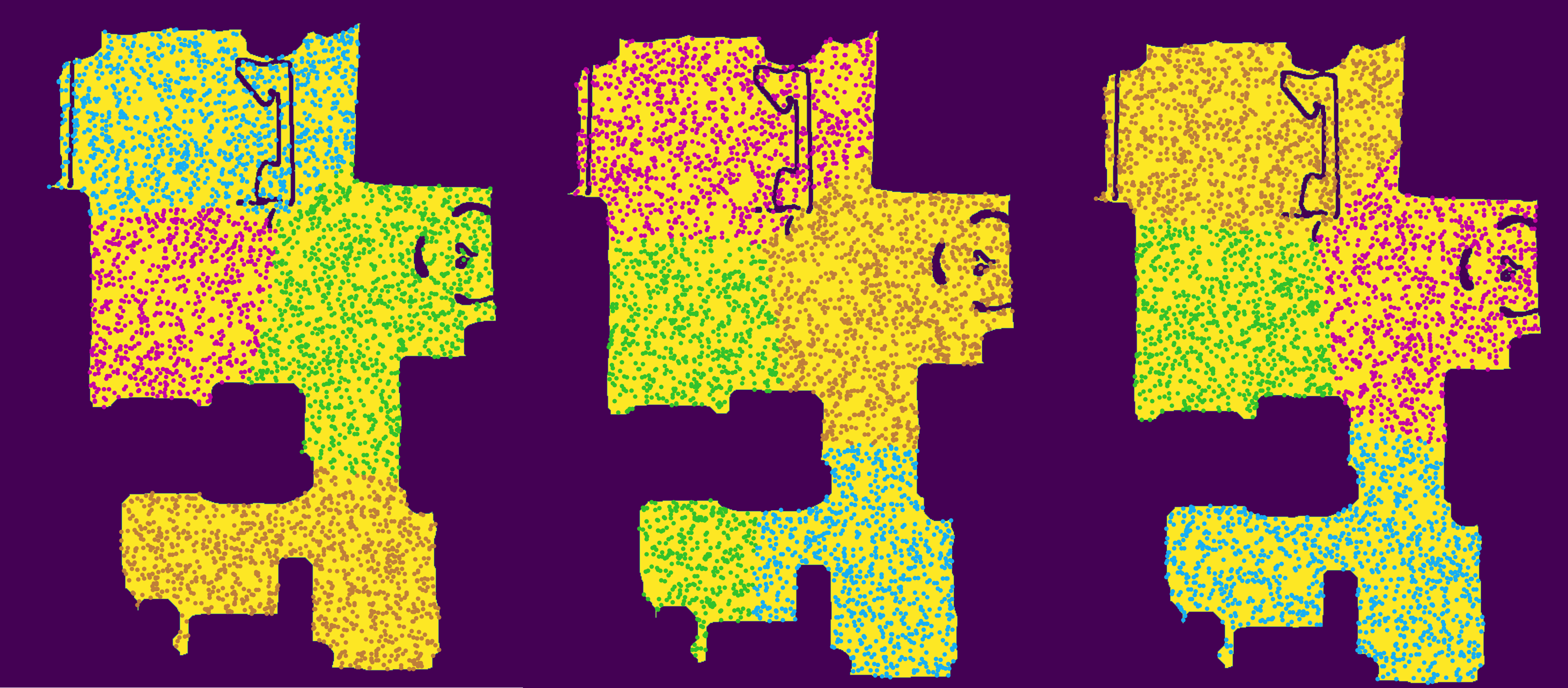}
\includegraphics[width=.7\linewidth]{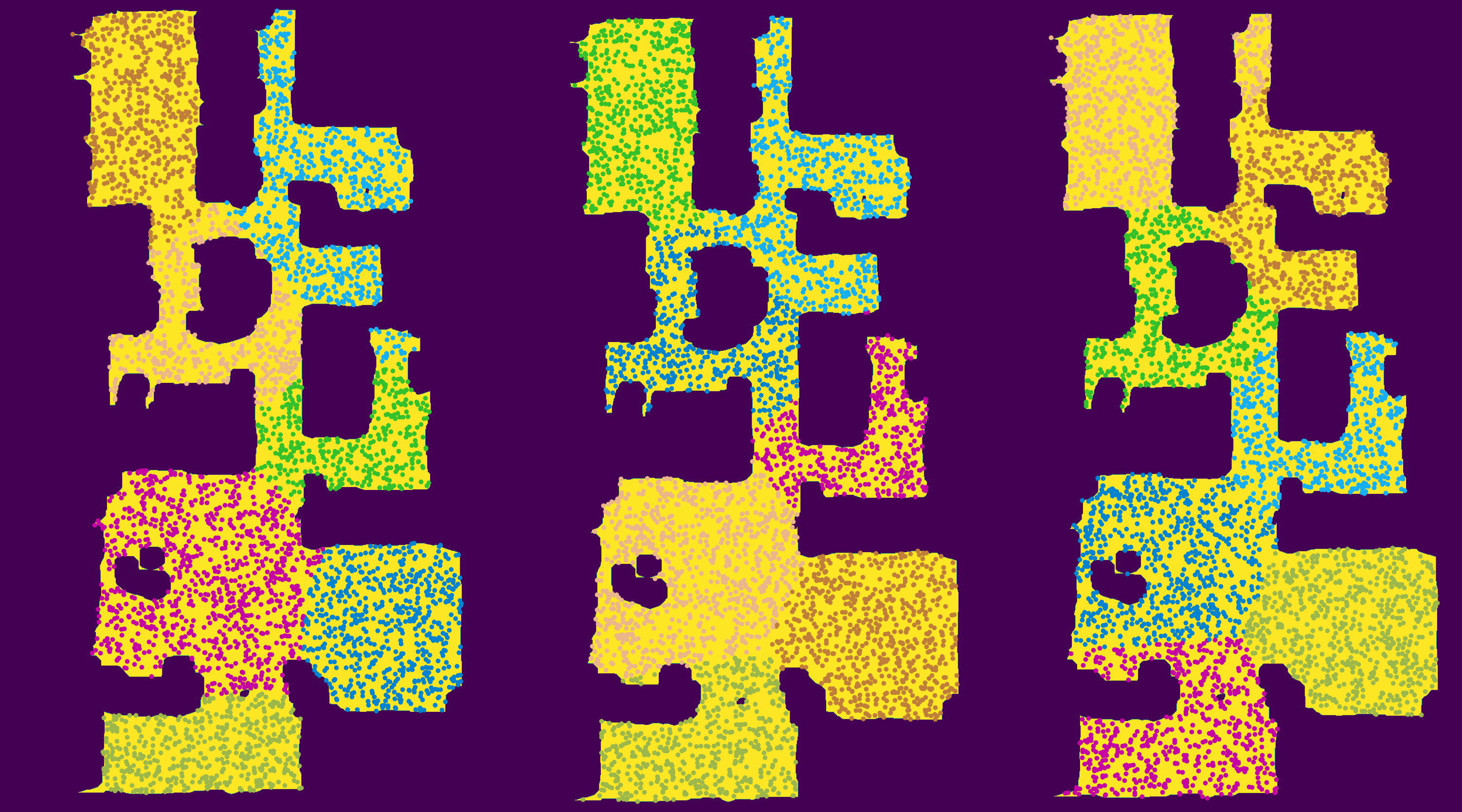}
\caption{Scene segmentation comparison, left to right (spectral, kmean, gmm), top to bottom ( Arbutus, Beach, Foyil)}
\label{fig:cluster}
\end{figure} 
 
GMM is more permissive regarding the shape of the cluster than k-mean so it captures rooms with diverse areas better. Qualitatively gmm also seems to offer the most reliable labelling thus making it the choice to automatically label our positions. The dataset creation procedure for one scene is the following:    

\begin{itemize}
  \item Load traversibility and obstacle maps to be able to discriminate valid positions ( at least 20 cm away from any obstacles/non traversable space) from non valid positions
  \item Per floor, sample 100 valid random positions and assign a room label to each position. This number is enough to capture relevant semantic cues within the floor even though it could be modified based on the SSA metric available in the metadata to maintain a constant point density.  
  \item For each coordinate position collect 54 RGB images. Images are collected in every direction with an increment of 20 degrees in azimuthal angle between two consecutive images. For each azimuthal angle 3 images are collected, one with camera looking downward, one with the camera looking straight, one with the camera looking upward.
\end{itemize}

Once this collection is over the splitting and preprocessing procedure is applied as follows for each polar angle:

\begin{itemize}
 \item Perform random filtering among the positions to have at most 10 positions per room and eliminate rooms with too few positions. The number of rooms decreases to 68 and 563 distinct positions are kept
 \item Perform random train/test split of remaining positions which yields 394 training positions and 169 testing positions
 \item For each training position add the 18 images associated with the selected polar angle to the training images with the corresponding room number as label. Images are normalized, converted to tensor and do not undergo augmentations.
 \item For each testing position add 6 (out of 18) randomly selected images associated with the selected polar angle to the testing images with the corresponding room number as label. Images are normalized, converted to tensor and do not undergo augmentations.
\end{itemize}

\subsubsection{Architecture and training}
\vspace{2mm}

For one polar angle, the final dataset is comprised of 7092 training images and 1014 testing images. The backbone of our classifier is a Resnet18 network pretrained on ImageNet1k (from pytorch) complemented with two fully connected layers of dimension 512 with leakyrelu activation and a fully connected output layer whose output dimension equals the number of rooms. Cross entropy loss with label smoothing to reduce overconfidence is used. Adam optimizer with a fixed learning rate of 1e-4 is sufficient. Training is performed during 15 epochs with a batch size of 20. Table \ref{tab:polar_room} shows the best classification metrics on the test set for different polar angles.  

\begin{table}[htbp]
\fontsize{10}{8}\selectfont
\centering
\caption{\bf Auxiliary task performances on test set}
\begin{tabular}{ccc}
\hline
Polar Angle & Accuracy & Macro F1\\
\hline
Upward & 0.8658 & 0.7916 \\
Neutral & 0.8392 & 0.7524 \\
Downward & 0.8500 & 0.7727 \\
\hline
\end{tabular}
  \label{tab:polar_room}
\end{table}

Camera looking upward yields the best performance because it captures more semantic information than the other orientations. 

\subsubsection{Image representation extraction}
\vspace{2mm}

\begin{figure}[htbp]
\centering
\includegraphics[width=.7\linewidth]{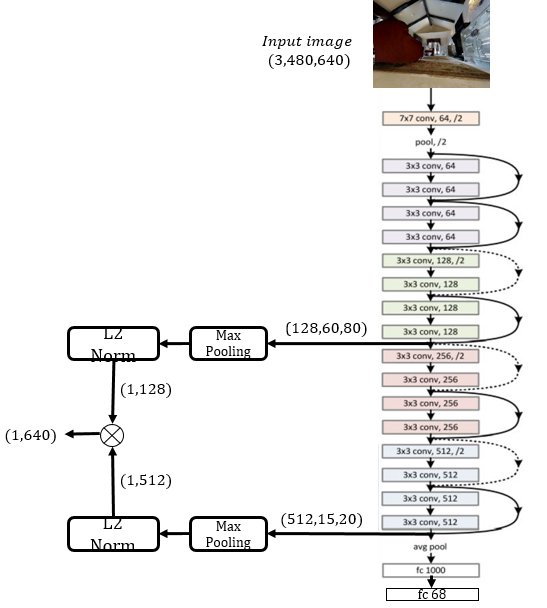}
\caption{Image embedding pipeline}
\label{fig:im_rep}
\end{figure}

Once this learning procedure is done we use a rather classical pipeline for image retrieval which consists in extracting local descriptors and pooling them in an orderless manner (\ref{fig:im_rep}). To extract local descriptors the finetuned Resnet18 backbone network is cropped at the 3rd and 5th layers. Cropping at multiple layers provide scale invariance while the descriptors extracted are robust to viewpoint change and slight lighting variations. The output shape of each layer is $H \times W \times D$ providing a $D$-dimension vector for each $H \times W$ location, maxpooling this output yields a $D$-dimensional representation that is l2-regularized afterward to obtain the final image representation. These operations also bring some robustness to translation and partial occlusion. 

\subsection{Local navigation}
\vspace{4mm}

The local policy's goal is to move forward from the direction the agent is facing while avoiding obstacles whether it is large structural elements or furniture. The local policy takes an embedded image corresponding to the agent's current visual sensors reading and output the action commands. Basing the decision on a single image may be brittle and lack robustness but the high frequency of readings and executed commands partially compensates any shortcomings on that side. We fix the decision frequency rate to 10Hz meaning that the local policy reads 10 images per seconds and output 10 commands per second while working with fixed length sequences of 5 seconds. This choice is arbitrary and seems like a good compromise between flexibility and grasping non trivial behavior. The high frequency combined with the complexity of the environments and the fact that the policy uses single images and not sequence of images implies that reactive policies would not be effective and some sort of memory is required. Thus we adopt a recurrent architecture despite its disadvantages for generalization. The command space is two dimensional with one value corresponding to linear velocity x while the second value corresponds to angular velocity z which is enough to fully control the agent. Reducing the command space dimension from 6 to 2 reduces the richness and diversity of behaviours but greatly facilitates the learning which is desirable.  

\subsubsection{Behavioral cloning}
\vspace{2mm}

As mentioned previously imitation learning is used to train the policy. In the learning from demonstration paradigm an expert operates the agent and provides demonstrations of the task that needs to be accomplished. Some observation/action pairs are therefore collected and this dataset is used to distill the expert knowledge into the local policy in a supervised manner. In real life collecting expert knowledge may be costly but this problem is alleviated in simulation. This behavioral cloning approach where the policy is learnt as a supervised problem over state-action pairs from expert trajectories is relatively simple but requires a lot of data to succeed due to compounding error caused by covariate shift because the model fits single timestep decisions. Minimizing a cross entropy loss assumes that the ground truth action distribution for a state is a delta function which is not the case if the latter is multi modal and high-dimensional. Thus the policy network can be fed a same input with different targets leading to high-variance gradients plaguing the learning. Therefore, following \cite{pathak2018zero} we learn a differentiable forward dynamic model $f_{\theta_{fm}}$ that predicts, in the feature space, from any state, the state the agent would end up to after a timestep if he was to take any action. A forward consistency loss is then derived. For a triplet consisting of a state in the feature space, the action taken and the resulting next state in the feature space $(u_t,a_t,u_{t+1})$ sampled from an expert trajectory; the consistency loss is a L2 loss in the feature space between the actual next state $u_{t+1}$ and the state predicted by the forward model when being fed the predicted next action and the actual state $\hat{u}_{t+1} = f_{\theta_{fm}}(u_t,\hat{a_t})$. Minimizing this loss with regard to the parameters of the local policy $\theta_{\pi}$ hence put the emphasis on reaching the goal independently of the action chosen. 

\begin{figure}[htbp]
\centering
\includegraphics[scale=0.5,width=.5\linewidth]{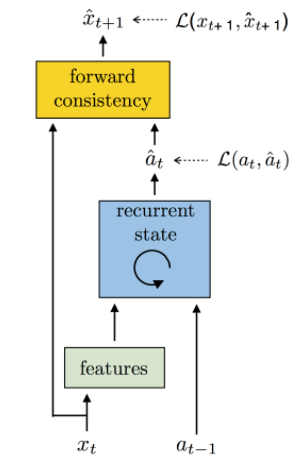}
\caption{Forward model structure, image from \cite{pathak2018zero}}
\label{fig:fm}
\end{figure}

\subsubsection{Collection of expert trajectories}
\vspace{2mm}

The simplicity of the approach combined to the accessibility of expert knowledge in simulation justify the behavioral cloning choice. ROS navigation stack is used to create that dataset of expert trajectories, it takes a starting and goal position within the scene and uses Lidar, RGBD sensors and odometry to output safe velocity commands to the robot. For each trajectory, the images, velocity commands as well as actual commands and odometry reading are recorded inside a rosbag. The local policy's behavior should be going forward while avoiding obstacles. It should not be initiating trajectories modification that would be not be motivated by obstacle avoidance. To constrain ROS navigation stack to create this specific kind of expert knowledge, we sample starting and goal positions from the edges of our topological maps generated on a subset of maps. Expert trajectories are thus collected on 20 scenes : Aloha, Arbutus, Beach, Bremerton, Cabon, Cason, Darnestown, Divide, Elton, Foyil, Frierson, Galatia, Gasburg, Kingdom, Mosinee, Natural, Newcomb, Rosser, Rutherford, Wyatt. In total 2486 expert trajectories are collected.

\subsubsection{Dataset creation}
\vspace{2mm}

Physics frequency as well as simulation frequency is 240Hz while the local policy works under 10Hz. This first task is to downsample to 10Hz the set of images in the rosbag. Then each of these images is associated to a velocity command from the same bag. The pair is matched by finding the command whose timestep is the closest to the image's own timestep. Velocity commands are only recorded when the latter changes, thus there is much less velocity commands than images in the bag. Afterward the trajectory is cut into multiple fixed length sequences of $L$ time steps. We choose to work with $L=50$ which accounts for 5s sequences at 10Hz, it is a good compromise to capture non trivial behaviors with limiting computational complexity. A maximum of 4s overlap is authorized between two sequences. Image are embedded by concatenating representations from the trained localization and passage detection networks as described in the image representation section, the resulting vector has a (1,1152) shape. Despite some loss of information we converted the problem to a classification task which is more stable and easier to train. The velocity commands were discretized into non overlapping uniform classes. Linear velocity in the range 0 to 0.5 was discretized in 3 classes while angular velocity in the range -0.5 to 0.5 was discretized in 5 classes. Finally straight trajectories are downsampled by eliminating any trajectory with less than 20 percent of non zero velocity commands. Doing so put the emphasis on obstacle avoidance and significantly improves navigation efficiency.

\subsubsection{Architecture and training}
\vspace{2mm}

The forward model is a MLP with an output dimension of 1152. It takes as input the current image's embedding an output the predicted next state. The local policy has one GRU cell dedicated to linear velocity with fully connected layers to output a distribution over the linear velocity classes and one GRU cell dedicated to angular velocity with with fully connected layers to output a distribution over angular velocity classes. Each cell takes as input the current embedded RGB image and the aggregated previous hidden states of the two cells. Having a cell for each dimension improves the performances compared to a single cell. However the two command dimensions are highly correlated so it is necessary to aggregate the hidden states of the cells to take into account that structural correlation. For a sequence, hidden states $h^{angular}$ and $h^{linear}$ are initialized to zero and hidden state size is set to 500. 

To stabilize the training, during the first 100 epochs the forward model $f_{\theta_{fm}}$ is pretrained by minimizing the forward consistency loss plus a regularization term while the policy network $\pi_{\theta_{\pi}}$ is frozen. Optimization is performed over the whole $L=50$ time steps sequence.  
\begin{equation}
\begin{tabular}{ll}    
$min_{\theta_{fm}}$&$ \sum_{t=1}^{L-1} ( \Vert u_{t+1}-\bar{u}_{t+1} \Vert_2^2 + \lambda \Vert u_{t+1}-\hat{u}_{t+1} \Vert_2^2 ),$  \\
%\vspace{2mm}
&where \\
&$\bar{u}_{t+1} = f_{\theta_{fm}}(u_t,a_t)$,\\ 
&$\hat{u}_{t+1} = f_{\theta_{fm}}(u_t,\hat{a_t})$,\\ 
&$\hat{a}_t = \pi_{\theta_{\pi}}(u_t,h^{angular}_t,h^{linear}_t)$,\\
\end{tabular}
\label{eq:optim1}
\end{equation}
where $u$ represents a state in the image feature space described earlier, $a$ an expert action and $\hat{a}$ an action predicted by the local policy $\pi_{\theta_{\pi}}$. For the rest of the 900 epochs, both the policy $\pi_{\theta_{\pi}}$ and the forward model $f_{\theta_{fm}}$ are jointly trained by minimizing the following loss with cross-entropy ($CE$) terms over sequences.

\begin{equation}
\begin{tabular}{ll}   
$min_{\theta_{fm},\theta_{\pi}}$&$\sum_{t=1}^{L-1} ( \alpha(CE(\hat{a}_{t,angular},a_{t,angular})$ \\
  &$+CE(\hat{a}_{t,linear},a_{t,linear}))$ \\
  &$+\Vert u_{t+1}-\bar{u}_{t+1} \Vert_2^2 + \lambda \Vert u_{t+1}-\hat{u}_{t+1} \Vert_2^2).$\\
&where \\
&$\bar{u}_{t+1} = f_{\theta_{fm}}(u_t,a_t)$,\\ 
&$\hat{u}_{t+1} = f_{\theta_{fm}}(u_t,\hat{a_t})$,\\
&$\hat{a}_t = \pi_{\theta_{\pi}}(u_t,h^{angular}_t,h^{linear}_t)$.\\
\end{tabular}
\label{eq:optim2}
\end{equation}

Hyper-parameters $\alpha$ and $\lambda$ control the prevalence of the cross-entropy loss compared to the forward consistency loss and vice-verca. Adam optimizer with a $1.e-4$ learning rate is used. Training is performed with batches of 50 sequences over 1000 epochs. The following table \ref{tab:locpol} presents classification performances averaged over each sequence on the test set for different hyper parameter values. 

\begin{table}[htbp]
\fontsize{10}{8}\selectfont
\centering
\caption{\bf Local policy training metrics on test set}
\begin{tabular}{cccc}
\hline
$\alpha$ & $\lambda$ & f1 & acc \\
\hline
\bf Angular velocity \\
\hline
0.1 & 0.5 & 0.41 & 0.47 \\
10 & 0.1 & 0.43 & 0.49 \\
50 & 1 & 0.42 & 0.48 \\
\hline
\bf Linear velocity \\
\hline
0.1 & 0.5 & 0.37 & 0.65 \\
10 & 0.1 & 0.39 & 0.63 \\
50 & 1 & 0.40 & 0.67 \\
\hline
\end{tabular}
  \label{tab:locpol}
\end{table}

The following figures show the output of the local policy during two navigation sequences and illustrate different behaviors. 

\begin{figure}[htbp]
\centering
\includegraphics[width=.7\linewidth]{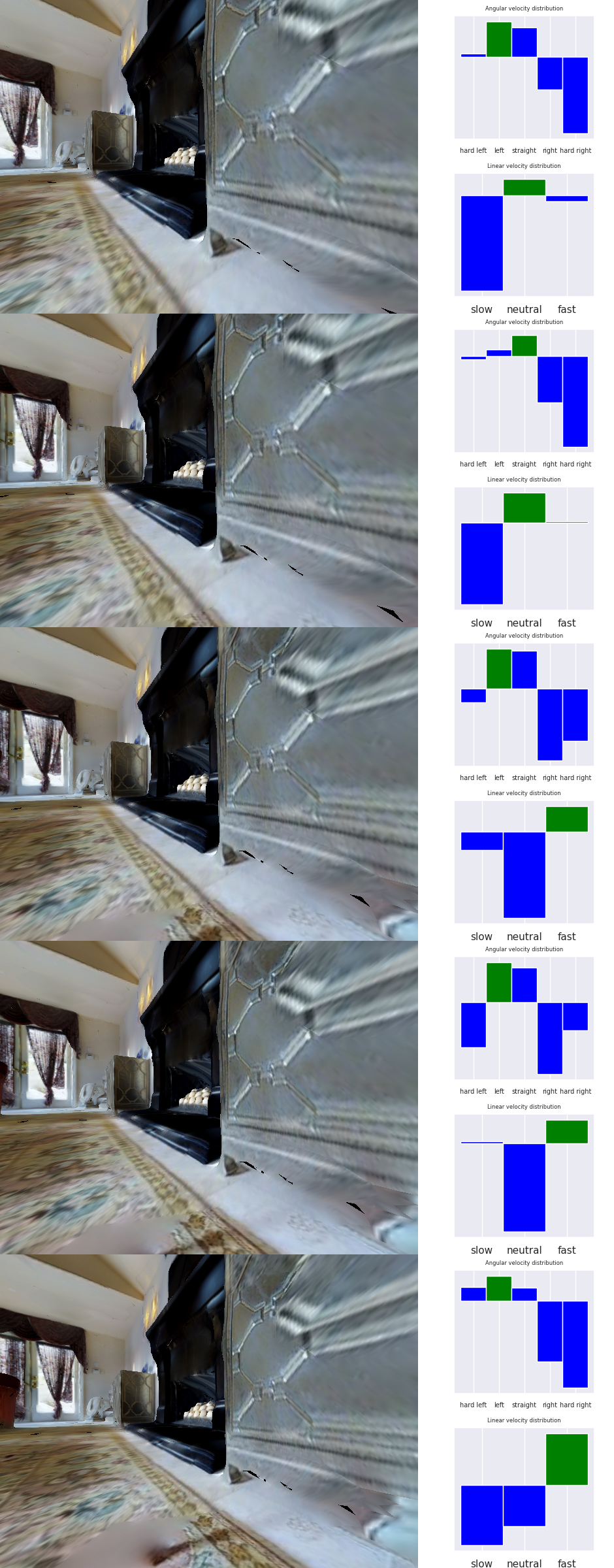}
\caption{Local policy behavior turn left (read bottom to up)}
\label{fig:navleft}
\end{figure}

Figure \ref{fig:navleft} shows the agent navigating toward a sideboard on its right. Angular velocity command distribution clearly have more mass on left turn commands with right turn commands having very low logits values. On the other hand linear velocity changes from fast to neutral as the agent approaches the obstacle. 

\begin{figure}[htbp]
\centering
\includegraphics[width=.7\linewidth]{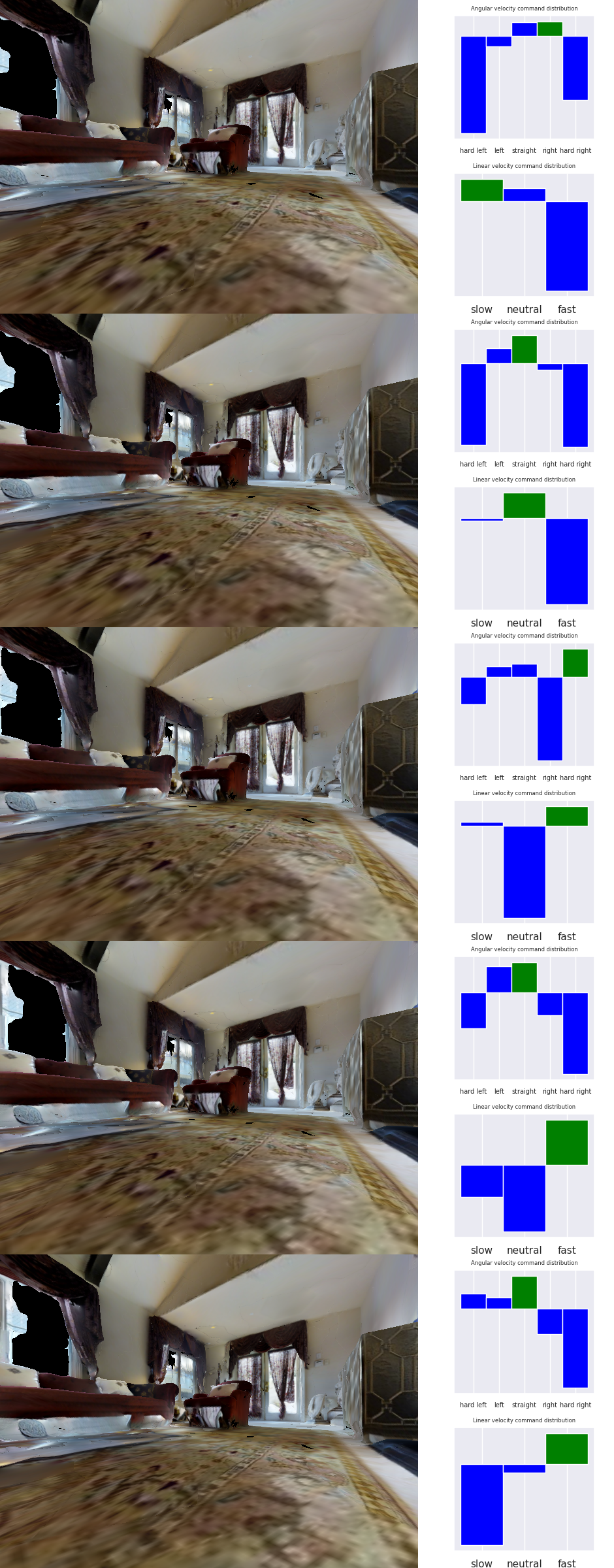}
\caption{Local policy behavior straight forward (read bottom to up)}
\label{fig:navs}
\end{figure}

Figure \ref{fig:navs} shows the agent navigating within a relatively obstacle free area. Fast linear velocity command are mainly sent by the local policy. Angular velocity command distribution are balanced with very low mass both turn commands. Very few turn commands are sent but for some correction toward the right.

\subsection{Passage detection model}
\vspace{4mm}

The passage detection model is a binary classifier used both in the topological map and the local navigation. It aims at implicitly capturing structural information about the scene such as doors, walls and obstacles. From a single RGB image, the model is trained at deciphering if the space in front of the agent is navigable. We create a labelled dataset and learn the model in a supervised manner. The latter should be robust to viewpoint changes and scale invariant while minimizing the false positives.  

\subsubsection{Dataset creation}
\vspace{2mm}

Still within the iGibson dataset, the scenes Elton, Gasburg, Arbutus, Natural, Rosser, Frierson, Wyatt are used to sample training images while the scenes Aloha, Beach, Foyil are used to sample testing images. For one scene, the data collection procedure consists in the following steps: 
\begin{itemize}
 \item Segment the scene into the appropriate number of rooms
 \item Per room sample 10 valid source positions (at least 30 cm away from the closest wall/obstacle) 
 \item Per source position draw a 2 meters radius circle having as center the source position. Sample 10 target positions on that circle covering every direction.
 \item For each source/target pair teleport the agent to the source position and orientate it toward the target position and save the image. Select 50 evenly spaced points on the source/target segment. Using the traversibility and obstacle maps, if at least one point of that segment falls within a obstacle or a wall assign label {\it non passage}, else assign label {\it passage}.
\end{itemize}

The dataset contains 7750 images and is unbalanced toward non passage image. Downsampling to a 1:1 ratio is necessary to improve the performances, thus the final dataset contains 4020 training images and 1234 testing images. Images are normalized and cropped before being fed to the model.

\subsubsection{Architecture and training}
\vspace{2mm}

The backbone of the classifier is a Resnet18 network pretrained on ImageNet1k (from pytorch) complemented with two fully connected layers of dimension 512 with leakyrelu activation and a fully connected output with one dimensional output. Using the pytorch pretrained Resnet18 rather than the Resnet18 trained on the room classification task yields better result. Training on the room classification is very specific and would create too much negative transfer for this task. Binary cross entropy loss is used combined with an Adam optimizer with a fixed learning rate of 1e-4. Training is performed during 20 epochs with a batch size of 32. Table \ref{tab:pass_d} shows the best classification metrics on the test set Resnet18 and Alexnet backbone networks.  

\begin{table}[htbp]
\fontsize{10}{8}\selectfont
\centering
\caption{\bf Passage detector performances on test set}
\begin{tabular}{ccc}
\hline
Backbone model & accuracy & f1 \\
\hline
Resnet18 & 0.8606 & 0.8558 \\
Alexnet & 0.8387 & 0.8418 \\
\hline
\end{tabular}
  \label{tab:pass_d}
\end{table}

\begin{figure}[htbp]
\centering
\includegraphics[width=0.95\linewidth]{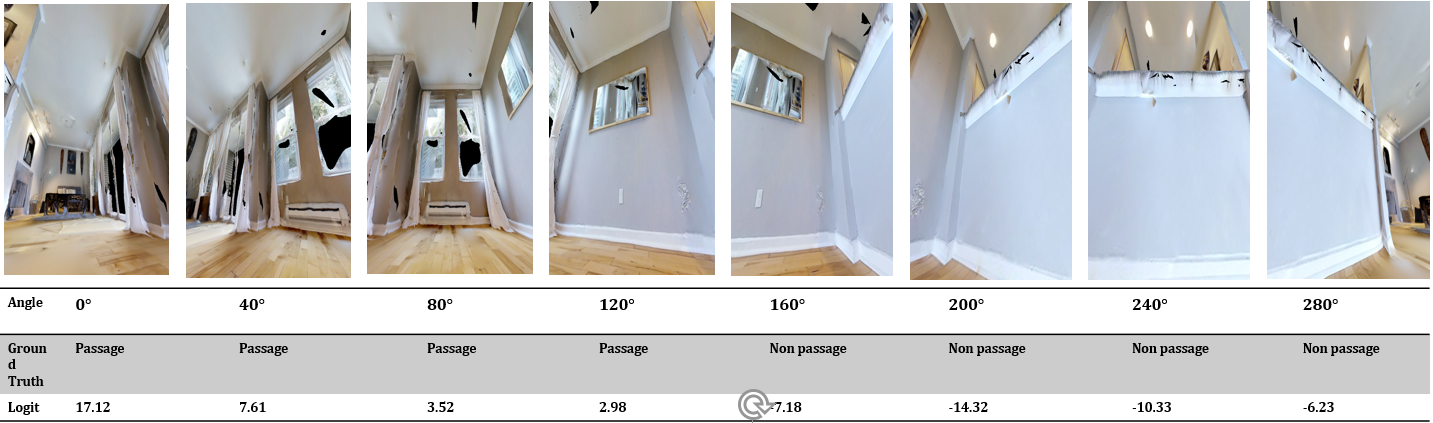}
\caption{Passage detector behavior}
\label{fig:pd1}
\centering
\includegraphics[width=0.95\linewidth]{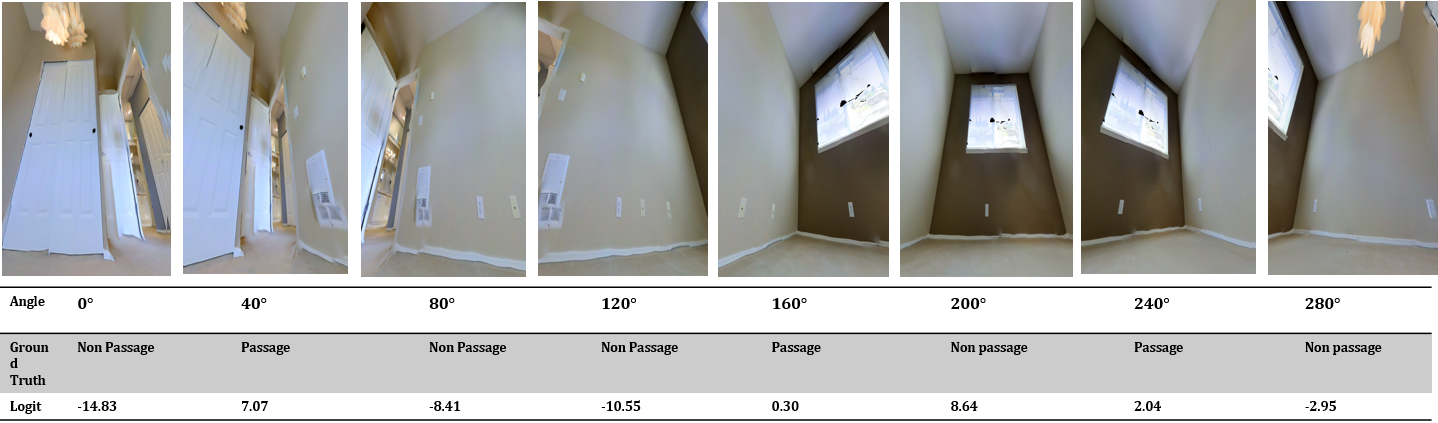}
\caption{Passage detector behavior}
\label{fig:pd2}
\end{figure}

As intended the final model is able to grasp the nature of its surroundings with decent accuracy. Errors fall within two categories; true errors whether they are false positives or false negatives and errors resulting from misguiding labelling. If a passage is in the image but not in the axis in front the robot the label will be non passage. When taking into account the logit values, there are even fewer images misclassified with high confidence. The main problem lies in the passage detector inability's to detect very thin obstacles such as chair/table legs. 
\ref{fig:pd1} and \ref{fig:pd2} illustrate the passage detector behavior when being fed images from a single position and different angles.

\vspace{40mm}
\subsection{Topological map}
\vspace{4mm}

The topological map is the key component in the navigation pipeline. It is a directed graph built as follows for a set of node positions: 

\begin{equation}
G_{topo map} = (V,E)\quad \rm{where}
\end{equation}
\begin{eqnarray*}
V & = & \{  EmbeddingModel(im_{node}) | node \in NodePositions\}\\
E & = & \{  -log(PassageDetector(im_{source\xrightarrow{}target}) \mid \\ 
  &   &     \quad (source,target) \in NodePositions^2,\\ 
  &   &     \quad d(source,target)<2.5m \}.
\end{eqnarray*}

\begin{figure}[htbp]
\centering
\includegraphics[width=1\linewidth]{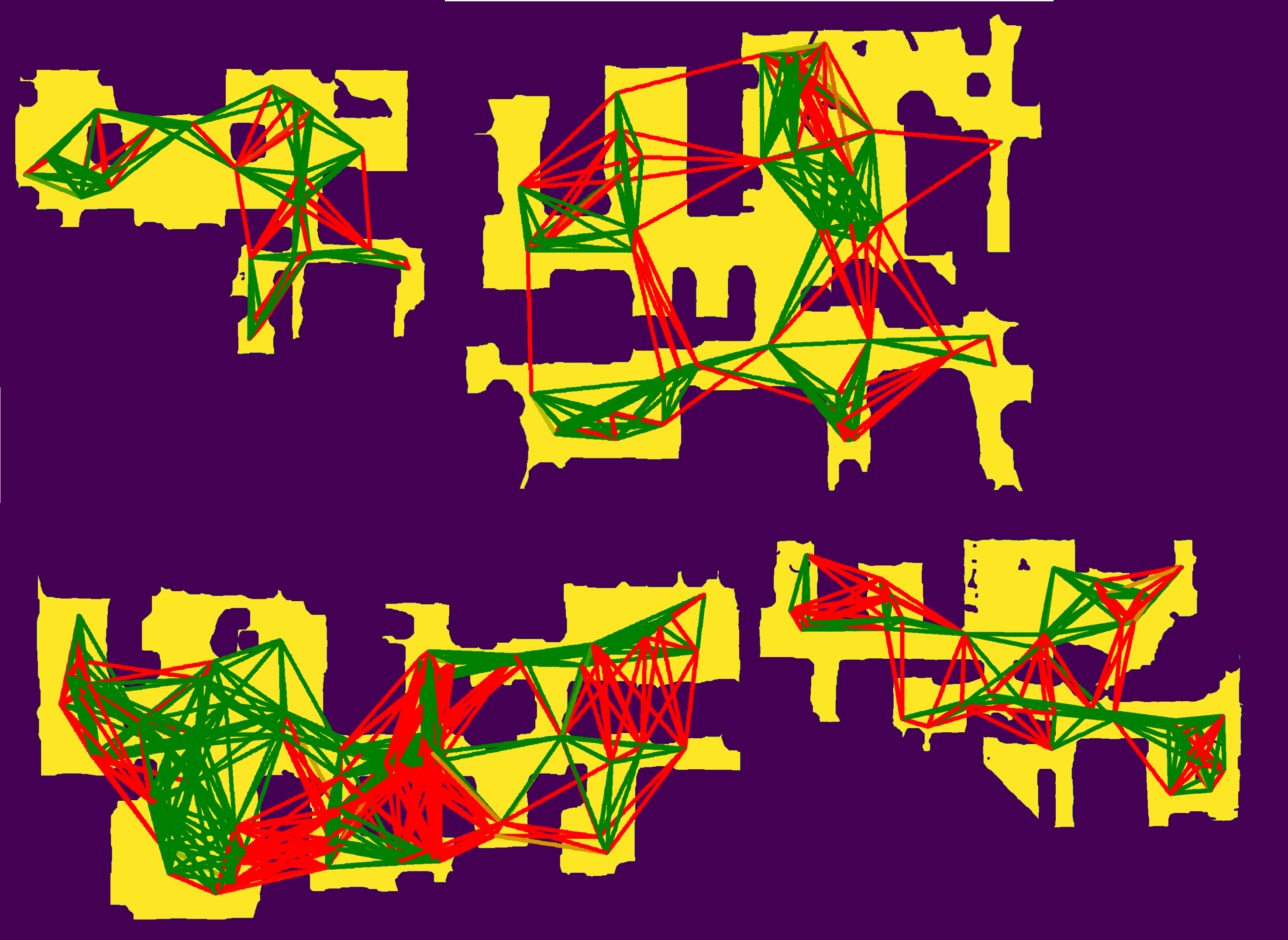}
\caption{Illustrated topological maps (Gasburg, Galatian Divide, Foyil)}
\label{fig:topomaps}
\end{figure}

A node corresponds to one position on the map and is represented as the embeddings of the images taken from that position orientated in 18 different angles covering a full circle (20° increment between two consecutive angles). An edge connects one source node to one target node with the log-likelihood of having a passage between the source and the target. The latter is computed by processing the image taken from the source orientated toward the target through the passage detector. The absolute angle between the source and the target is also stored. The perspective between source/target and target/source being different it is necessary to have directed edges. For example if there is no passage between two nodes, there may be more navigable space before hitting the obstacle on one perspective than on the other. Figure \ref{fig:topomaps} shows some topological maps generated for a few maps. The greener the edge the higher the probability of having a passage, the redder the lower the probability of having a passage.

The graph is not complete, only nodes within a close vicinity are connected to maintain a somehow realistic layout. However the graph must be connected to ensure that each part of the scene is reachable. The topological map must maintain the best scene coverage as possible to ensure a smooth navigation. Therefore, the number of nodes is a crucial parameter that influences the navigation's quality. The lower the number of nodes, the higher the distance between the robot actual position and the node position thus the less reliable the ensuing angle indication is. Some local goals may also become less reachable by lack of better choice. However the higher the number of nodes, the higher the probability of wrong localization by having very similar node images. Trajectories also tend to become non optimal from a metric standpoint because planning is based on navigability. A reasonable comprise has to be found.        

As explained in the problem setup any exploration phase where the topological map could be built is excluded. Thus, in practice the nodes were selected by sampling random positions and ensuring a minimum position density per room. This method is certainly less optimal than organically building a map throughout an exploration phase because nodes are chosen randomly and do not correspond to spaces where the robot has intuitively navigated throughout the exploration phase.

\section{Navigation evaluation}
Now that all the modules are designed and trained it is necessary to evaluate them. The example subsection will demonstrate one successful navigation episode and one failed navigation episode to illustrate the agent's behavior and underline recurrent characteristics about the pipeline. The other subsection provides an evaluation protocol and some evaluation metrics on a more significant number of episodes. As explained before the agent is controlled through ROS, the iGibson simulator/scene being instancied as a node and the python navigation script as another node.
\subsection{Examples}
\begin{figure}[htbp]
\centering
\includegraphics[width=0.7\linewidth]{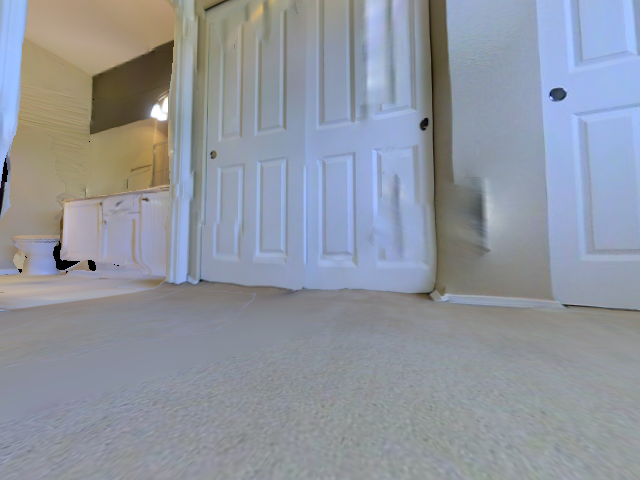}
\caption{Goal Image episode 1}
\label{fig:goal1}
\includegraphics[width=0.7\linewidth]{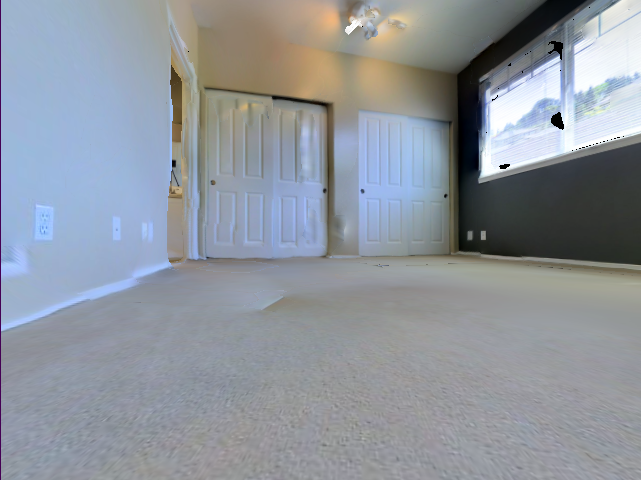}
\caption{Closest image found episode 1}
\label{fig:infgoal1}
\end{figure}

\vspace{4mm}
The following examples take place on the Aloha scene which has a relatively complex layout with doors and corridors but is not furnished. This is ideal to fully grasp the agent's behavior. The agent is teleported to a random position. It is given a goal image and must reach it within 30 minutes in the OS time referential.
\ref{fig:goal1} shows the goal given to the agent for the first navigation example. Its starting position is located 10.7 meters away from the goal while the shortest path measures 13.3 meters. The goal is reached in 152 seconds in simulation time which accounted for 26 min in the OS time. The closest image to the goal found throughout the navigation is displayed in \ref{fig:infgoal1}, the latter triggered the end of the episode due its similarity with the actual goal.

\begin{figure}[htbp]
\centering
\includegraphics[width=0.7\linewidth]{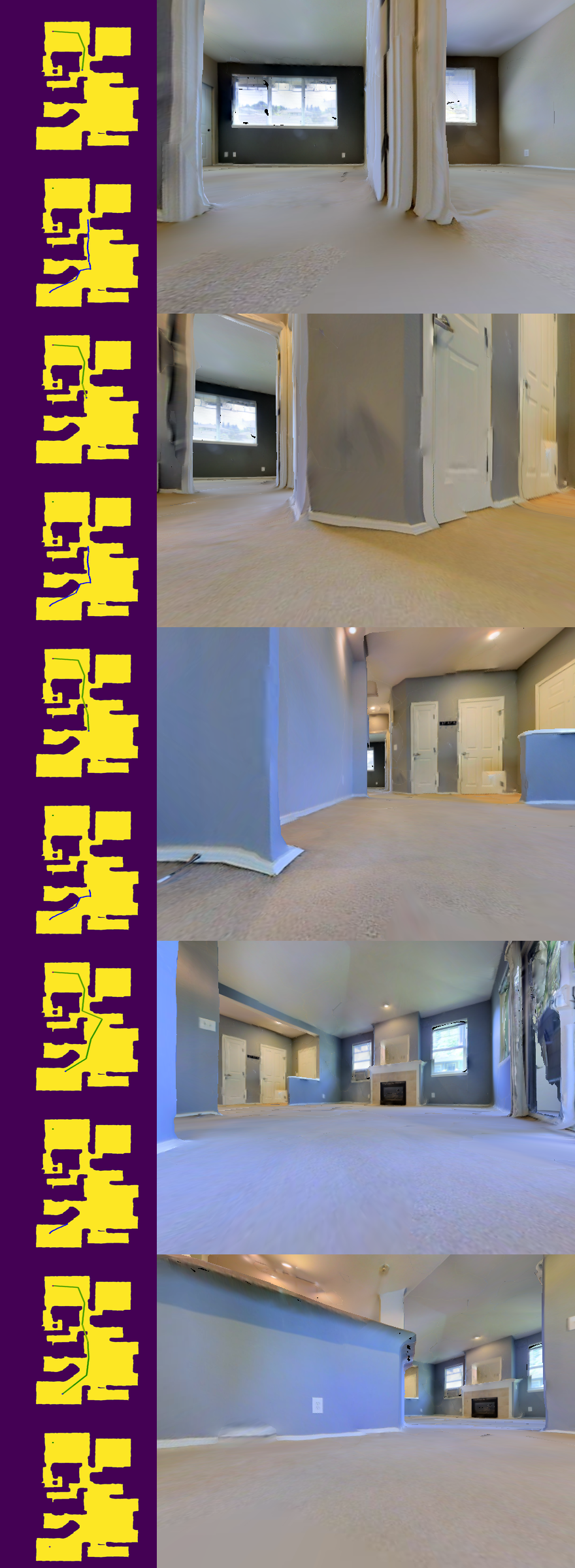}
\caption{First navigation episode (read bottom up)}
\label{fig:nav1}
\end{figure}

The first episode was successful for multiple reasons. Localization has been accurate throughout the whole episode allowing relevant planning and effective replanning as soon as localization information was updated. It is observable on the top left map of \ref{fig:nav1}. The topological map did provide a good scene coverage with a reasonable node density which resulted in realistic, safe and navigable trajectories. Finally local policy acted as expected, moving the agent forward while performing slight trajectory modifications to avoid obstacles. This happens on the frames 3,5 of \ref{fig:nav1} while the passage detector explicitly modified the robot orientation on frame 4.     
\vspace{5mm}

\begin{figure}[htbp]
\centering
\includegraphics[width=0.7\linewidth]{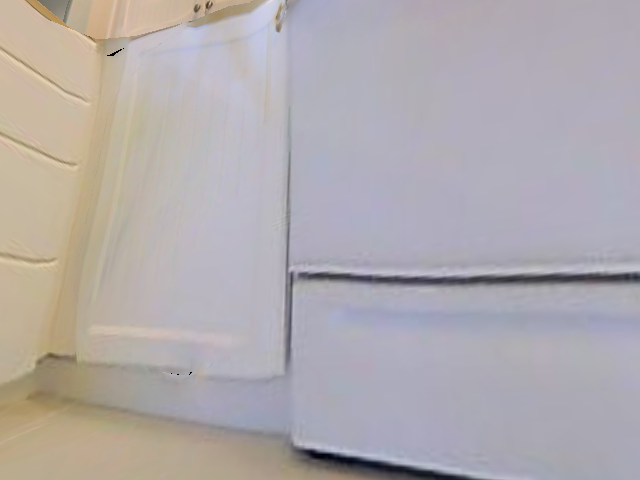}
\caption{Goal Image episode 2 }
\label{fig:goal2}
\end{figure}

\ref{fig:goal2} shows the goal image provided for the second navigation episode. Starting location remains the same. The latter is 3.56 meters away from the goal with a 7.34 meters shortest path. Robot collided with a wall after 55 seconds of simulation or 9 minutes in cpu time resulting in run termination.

\begin{figure}[htbp]
\centering
\includegraphics[width=0.7\linewidth]{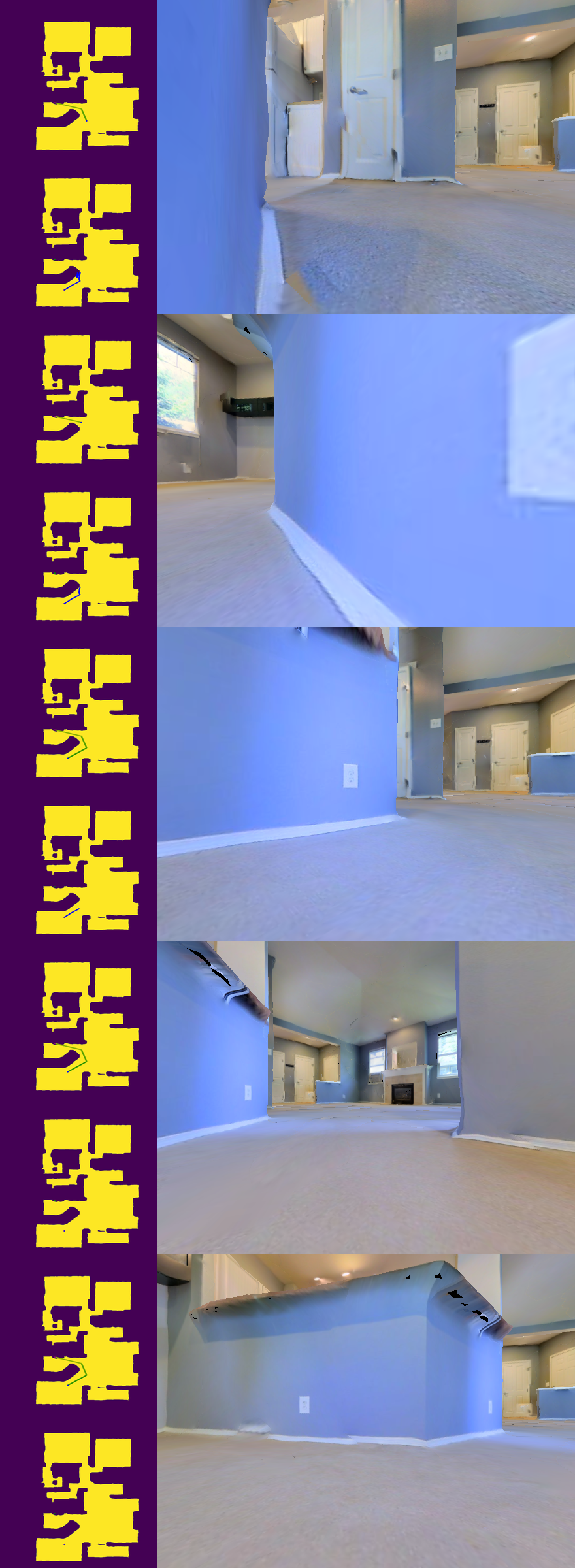}
\caption{Second navigation episode (read bottom up)}
\label{fig:nav2}
\end{figure}

Two issues that broadly apply to the pipeline can be underlined. The localization is discrete meaning that there will always be some distance between the robot actual position and the node position even if the localization if successful. The angle to reach the next goal node was computed from the node position and may not be suited to the agent actual position especially in tiny areas and passages. This structural limitation and underlying compromise was addressed in the topological map subsection. On \ref{fig:nav2} frame 3, the agent keeps orientating itself toward the blue wall because the planning module communicates an angle that would have been correct had the agent been 1 meter further (not a localization nor planning error in itself). On the other hand the local policy and passage detector successfully act to avoid collision (frame 4) but move the agent away from the global goal. The second issue lies in the fact that the robot base is much larger than the camera. Therefore some collisions may happen even though the visual input suggest that the agent dodged the obstacle. This happens on frame 5 of \ref{fig:nav2}.   

\subsection{Quantitative evaluation}
\vspace{4mm}

To gain more insight on the navigation pipeline's behavior, a thorougher evaluation is required. To that end we select seven maps, Aloha, Beach, Divide, Foyil, Galatia, Gasburg and Sasakwa. This is a little subset of maps due to the fact that that the pipeline was trained on a small subset of maps in the first place. Sasakwa is the only scene not present in at least one training dataset. The scenes present a wide variety of layouts and furniture density, thus the pipeline will be tested on multiple navigational complexity levels.

\vspace{3mm}
Depending on the scene's surface twenty-five to sixty episodes are randomly sampled. An episode consists in a random starting position and orientation where the agent will be teleported to at the beginning of the episode, a random goal image that the agent will receive at the beginning of the episode and the shortest path distance used to compute some evaluation metrics (computed from metadata). To ensure that the sampled episodes cover all the navigation paths that the scene may offer, we segment the scene and ensure that starting and goal positions belong to different clusters. Furthermore trivial and low complexity episodes are discarded by deleting episodes where start and goal positions would be located less than 2.5 meters away. Each episode is independent, the agent does not learn nor store information between two episodes. The agent is always provided a topological map of the associated scene and the trained models of each module.

\vspace{3mm}
Five evaluation metrics are used; success rate (SR), success weighted by path length (SPL), the rate of successful runs with at least one contact (RC), average aggregated duration of contacts during the successful episodes (AADC), average duration of the episodes (AD). SPL gives an indication of the optimality of the path taken by agent. It is the success rate weighted by the fraction of shortest path length over navigated path length. Os time referential is used to measure time.

\vspace{3mm}
Physical contacts are detected on the iGibson simulator and published on a specific topic. The navigation node subscribes to that topic to retrieve related information. Contact information is updated at a 10Hz rate. If there is at least one contact during a 5s local policy phase 5 seconds of contact time are added to the metric AADC. Within iGibson a contact is defined through pybullet physics simulator. At each step the latter records the identification number of any object that collides with any part of the agent. These ids are filtered to discard any contact between two parts of the agent and contact between the agent and the artificial floor. The remaining contacts occurred between the agent and the scene and are the one being monitored. In iGibson the floor is an artificial perfectly flat surface that is added on top of the original scene's floor. Thus some contact between the agent and the scene may be recorded whilst they are not 'penalizing' contact with walls or furniture for example.  

\vspace{3mm}
An episode is considered successful if the agent localizes itself in the same node as the goal node or if the agent finishes a local navigation episode within 1.5 meters of the goal. The run is not stopped if the agent collides because the contact definition is very sensitive and because it is more informative to evaluate the navigation quality.
\begin{table}[htbp]
\fontsize{9}{7}\selectfont
\centering
\caption{\bf Navigation evaluation}
\begin{tabular}{cccccc}
\hline
Scene & SR & SPL & RC & AADC(s) & AD(s) \\
\hline
Aloha & 0.87 & 0.65 & 0.57 & 63 & 561 \\
Beach & 0.96 & 0.78 & 0.51 & 60 & 374 \\
Divide & 0.56 & 0.45 & 0.78 & 36 & 430 \\
Foyil & 0.52 & 0.43 & 0.77 & 93 & 644 \\
Galatia & 0.75 & 0.56 & 0.76 & 96 & 720 \\
Gasburg & 0.84 & 0.67 & 0.63 & 82 & 538 \\
Sasakwa & 0.89 & 0.73 & 0.17 & 27 & 293 \\
\hline
\end{tabular}
  \label{tab:nav_ev}
\end{table}

\vspace{3mm}
As expected, the efficiency of the navigation directly depends on the complexity of the environment. On uncluttered scenes or scenes with favorable layouts (Beach, Aloha) the performances are satisfactory with high SR, SPL and a majority of successful runs without any collisions. On more complicated scenes such as Gasburg, Galatia performances slightly decreases while on very complicated scenes such as Foyil or Divide performances drop to a 50\% success rate with rather high collision numbers. Metrics on Sasakwa are competitive which suggests that the pipeline has a good generalization capability when confronted to fully unknown iGibson scene. 

\begin{figure}[htbp]
\centering
\includegraphics[width=0.7\linewidth]{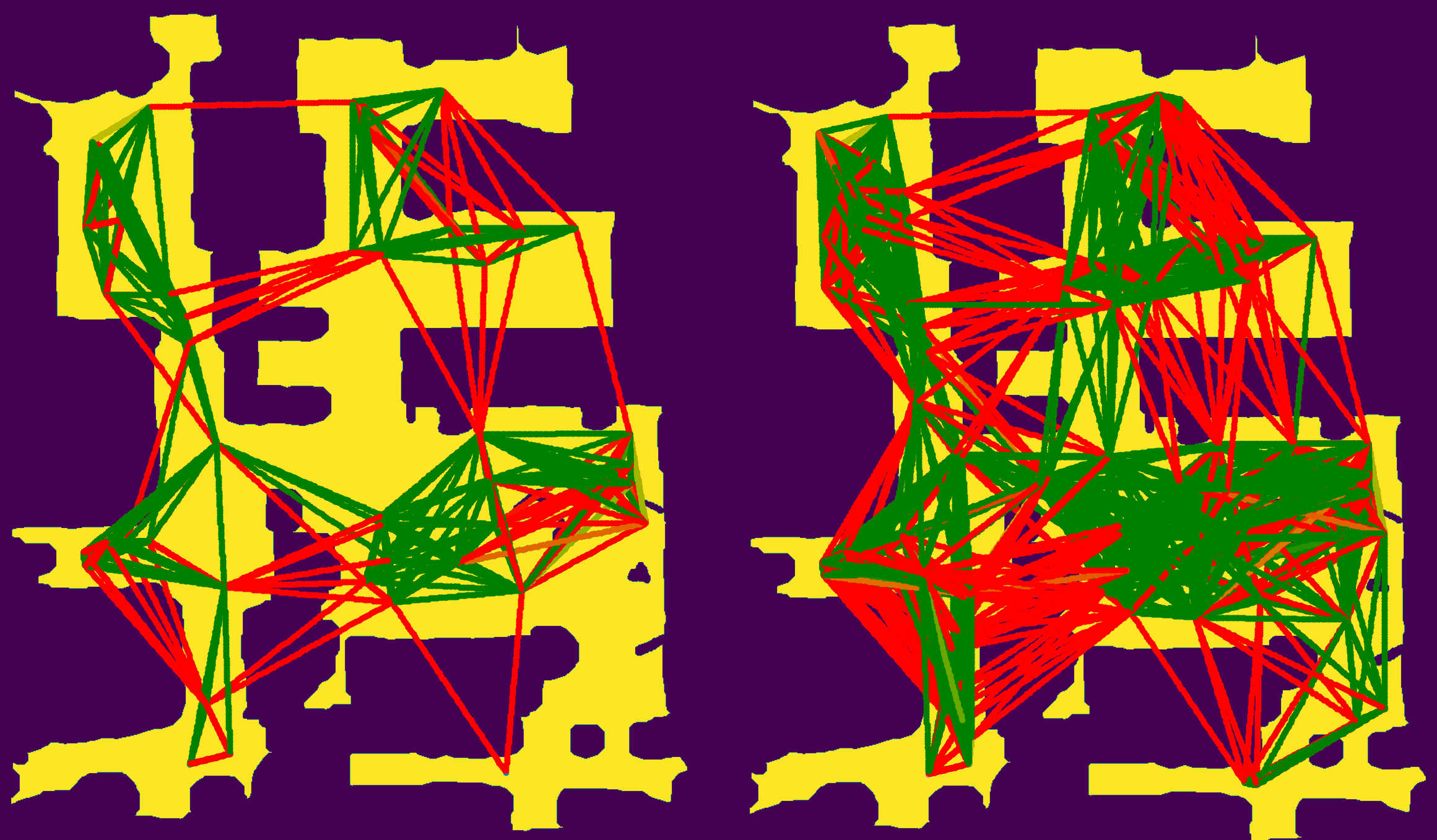}
\caption{Galatia maps comparison}
\label{fig:maps_diff}
\end{figure}

Overall, critical failures and collisions can be attributed to two main factors; faulty or non adapted topological maps and local policy failing to avoid obstacles. To illustrate the importance of having a topological map enhancing navigation, results for the same set of episodes are provided  in \ref{tab:maps_diff} but from two different topological maps (\ref{fig:maps_diff}). The first graph is not connected while the second graph fully covers the scene with an adequate density.

\begin{table}[htbp]
\fontsize{8}{7}\selectfont
\centering
\caption{\bf Performances rely on map quality}
\begin{tabular}{cccccc}
\hline
Scene & SR & SPL & RC & AADC(s) & AD(s) \\
\hline
Galatia bad& 0.55 & 0.36 & 0.87 & 110 & 817 \\
Galatia good& 0.75 & 0.56 & 0.76 & 96 & 720 \\

\hline
\end{tabular}
  \label{tab:maps_diff}
\end{table}

\subsection{Extensions}
\vspace{4mm}

The discrete nature of the topological map has consequences on the localization accuracy and trajectory feasibility. The structural limitations of the topological map within this navigation pipeline could be partially alleviated if the latter was built throughout an exploration phase. This however would change the physiognomy of the task and may lead to major modifications in the pipeline.
Improving this navigation pipeline without modifying the framework could be done in one of the following ways. 
\begin{itemize}
    \item Applying the dataset collection procedures on the full set of iGibson scenes and retraining the models on those newly acquired dataset. All the models would gain in robustness and performances.
    \item Using a self-supervised framework for learning image representation 
    \item Mapping with Djikstra's algorithm on sometimes lead to counter-intuitive trajectories where the optimal path is discarded in favor a of longer path whose images are characterized by deeper spatial perspectives. Mapping from a rightly defined constrained optimization problem on the graph may be beneficial.
    \item Changing the rather simplistic behavioral cloning approach for some inverse reinforcement learning approach. Or complementing the local policy with a finer way of detecting obstacles especially thin furniture.
\end{itemize}

\section{Sim2real}
The modularity of the pipeline allows us decompose the full transfer into multiple individual transfer tasks that are easier to tackle. Namely the following sections will independently describe the transfer of the passage detector, image representation extractor as well as the topological map.  
\subsection{Layout}
\vspace{4mm}

The real world navigation scene we picked is the second floor of the NLE's castle. The floor contains a big central room crossed by a spiral staircase in its center. On large sides, the room is bordered by offices with wooden doors and glass wall with white shade. A small wall is flanked by a door leading to a corridor while the opposite small wall either leads to a conference room or an utility zone through doors. The conference room is illuminated by 4 windows bringing a lot of natural light. The corridor is painted in white bordered by offices with white doors. Some office furniture lies in the central room and conference room. The central room and corridor are only illuminated by artificial lighting while the offices may be illuminated by artificial lighting or a unique small window.

\begin{figure}[htbp]
\centering
\includegraphics[width=0.7\linewidth]{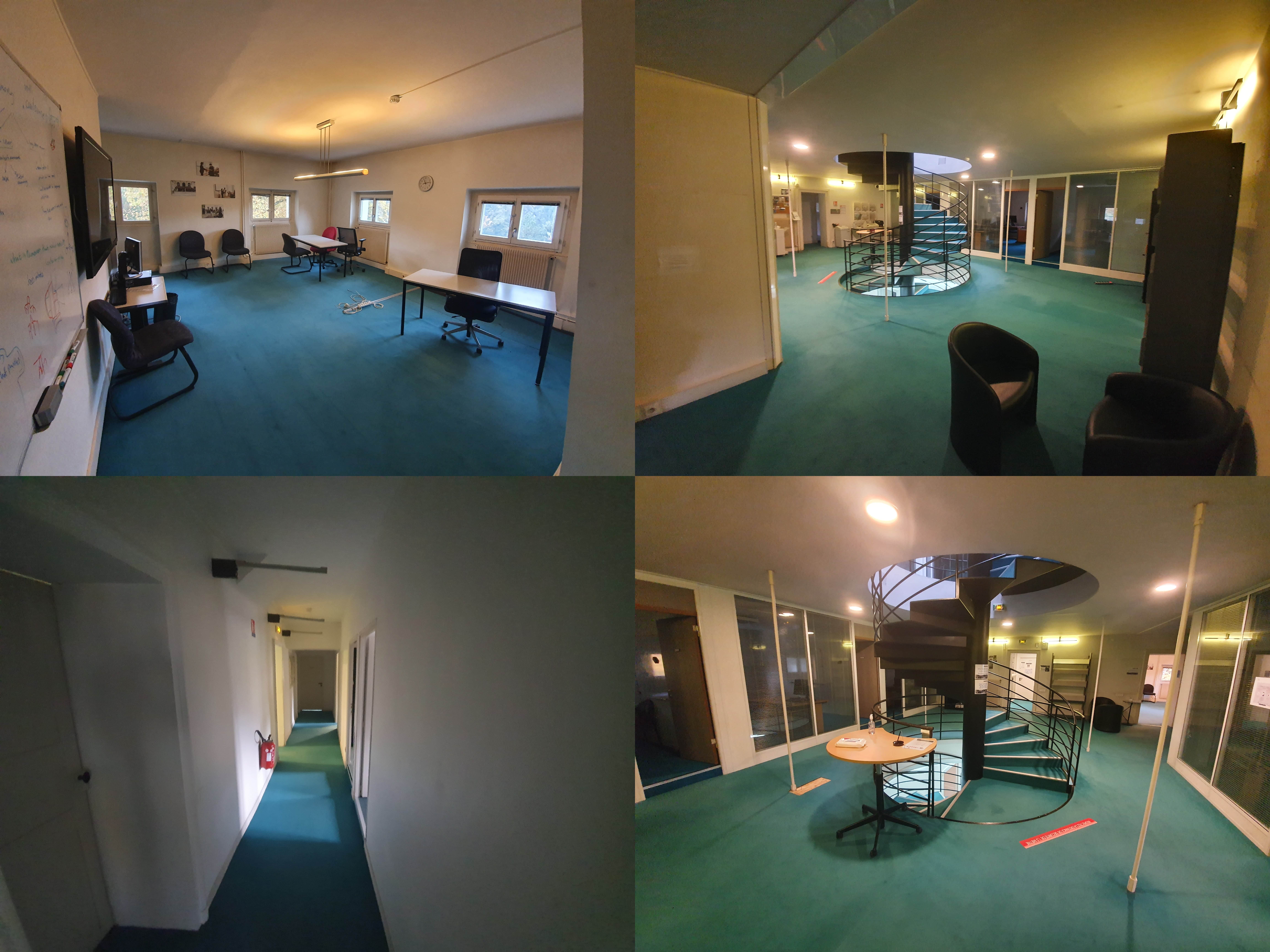}
\caption{Real navigation scene (source : phone camera)}
\label{fig:2floor}
\end{figure}

\vspace{3mm}
This is a challenging environment for multiple reasons. Contrarily to the wide majority of iGibson scenes, our scene has very few distinctive features, the semantic space is poor with very few cues. The walls are all painted with the same white paint and the floor is uniformly covered by a turquoise carpet. Lighting is a mixture of natural light and neon light, thus illumination field is very inconsistent and varies depending on the position and weather. The central room has four structural support white vertical bars (visible on \ref{fig:2floor}) that may hamper navigation due to their lack of detectability. Furthermore the floor is not perfectly smooth and flat with some light slopes or uneven transition to cross doors.  

\subsection{Dataset collection}
\vspace{4mm}

The first step toward transfer consists in collecting enough data from the scene. Collected data will be used both to transfer pipeline models and create a suited topological map. Gathering data is time-expensive, manually labelling it even more. Thus this process has to be done efficiently with a specific purpose in mind.  
Therefore we chose to primarily collect images to finetune the passage detection model and create the topological map. Other tasks require less specific images so that the collected images can be reused without any necessity to collect other images. 

\begin{figure}[htbp]
\centering
\includegraphics[width=0.7\linewidth]{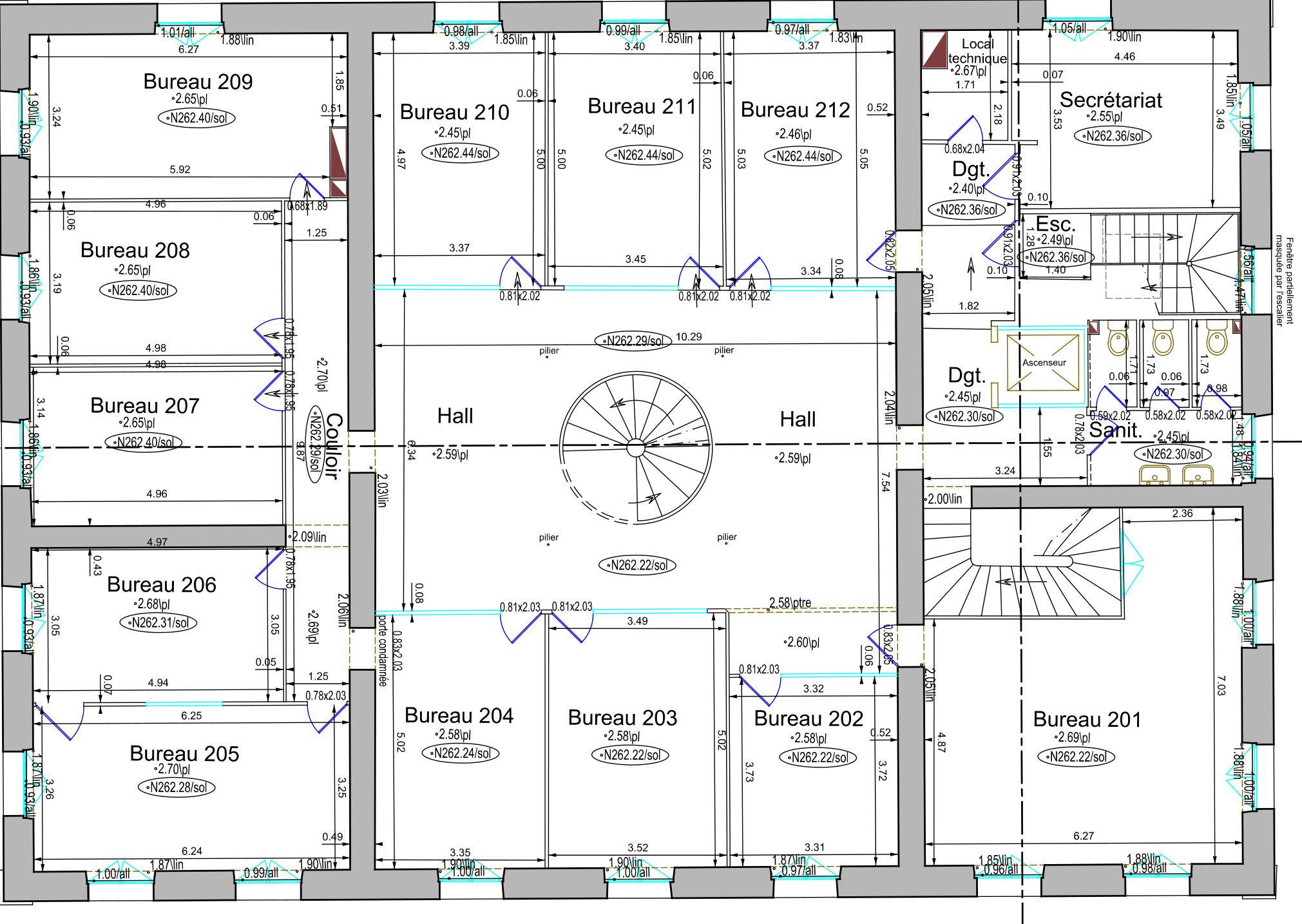}
\caption{Architectural map}
\label{fig:plan_raw}
\centering
\includegraphics[width=1\linewidth]{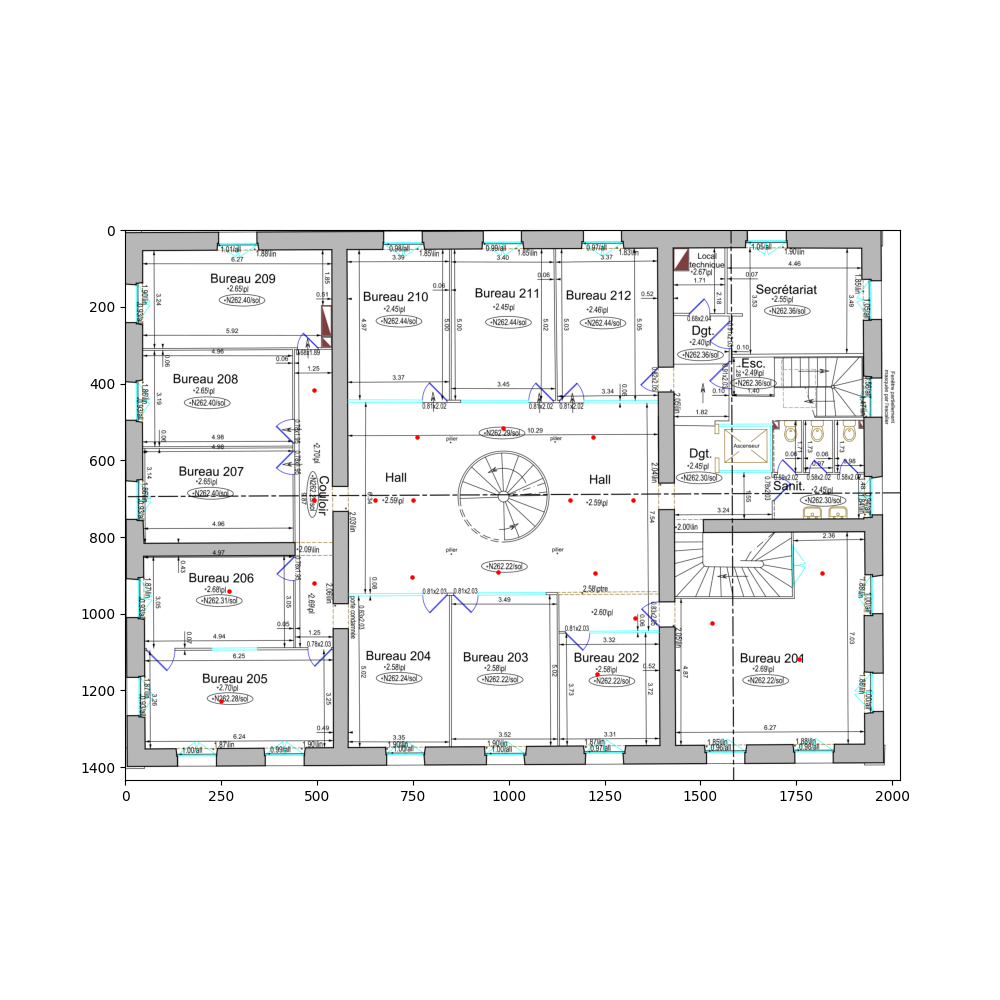}
\caption{Selected points}
\label{fig:plan_points_real}
\end{figure}

\vspace{3mm}
We recovered an architectural map of the scene with a proper scale as shown in \ref{fig:plan_raw}. From that map the collection procedure is performed as follows

\begin{itemize}
    \item Manually pick a set of strategic points. The set of points must fully cover the scene, be located at crossroad positions that enable navigation and must include a balanced proportion of images with navigable space/passages. The set of points are shown in red on \ref{fig:plan_points_real}, twenty positions are selected. Some offices and areas are not covered because they are separated from the rest of the scenes by a insurmountable step for the agent.
    \item On the map, distances are measured and converted to real metric distances to be able to position each selected point on the real scene. 
    \item Assign a unique numeric label to each point
    \item For each point, manually move the TTbot to its associated real location and start the image collecting procedure. The image collecting procedure captures an image, commands the agent to rotate from twenty degrees and repeats until the full circle is covered. Thus one position is associated with 18 images with different angle covering every direction. An image is uniquely designated with its position label and angle.
\end{itemize}

To comply with the navigation pipeline assumptions we must maintain a reliable and unified angle estimation source. In practice, we use visual odometry from the mounted RealSense camera to provide angle estimation. The latter provides angle differences from a reference direction, this reference direction corresponds to the direction the agent is facing when the camera is initialized during the roslaunch. We set the geographical North as this reference direction, thus the TTbot must always be initialized facing North. From all empirical evidence this angle estimation source is reliable even after long utilization phases. The image collecting procedure is executed on the TTbot's embedded ubuntu operating system.

\vspace{3mm}
To complete the passage detection real dataset, the 360 collected images must be labelled. In the first place six labels that identify the passage's position on the image are created; no passage, passage in front, passage left, passage right, passage extreme left, passage extreme right. This is estimated manually be individually looking at the image. Contrarily to the synthetic passage detection dataset, images with passages not directly in front of the agent are labelled as passage, this more lenient labelling aims at creating less confusion.

\begin{figure}[htbp]
\centering
\includegraphics[width=0.7\linewidth]{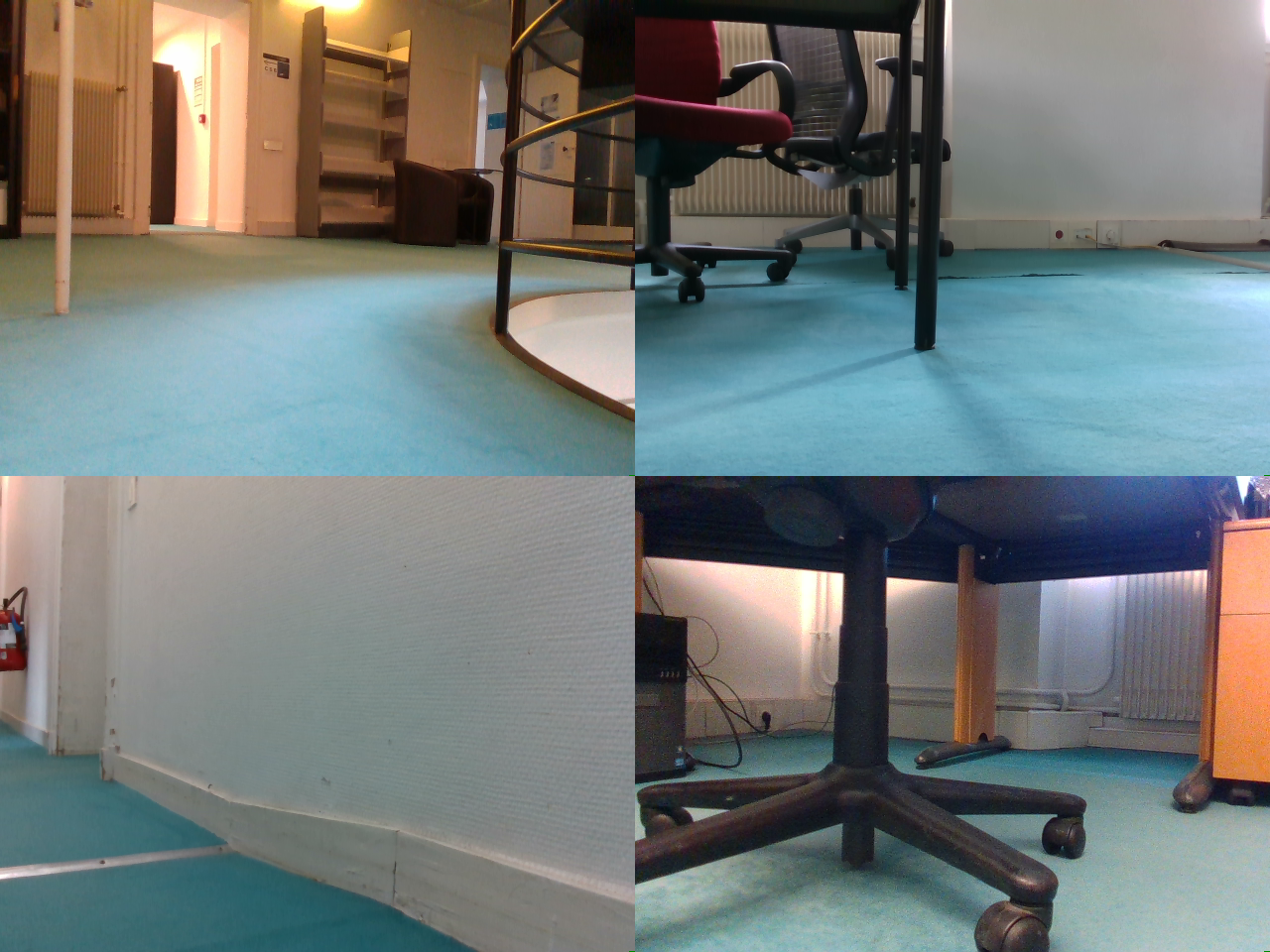}
\caption{Passage images, labels from left to right, top to bottom passage: right passage, extreme left passage, no passage}
\label{fig:plan_raw1}
\end{figure}

\subsection{Feature adaptation}
\vspace{4mm}

Domain adaptation is applied to learn a new image representation extractor network. Given an image from iGibson simulator $im_s$ sampled from $IM_s$, the image representation extractor trained in simulation $f_s$ output a feature vector $x_s = f_s(im_s) \sim X_s$ which follows the simulation feature distribution $X_s$. Given a real image sampled from real world image distribution $im_t\sim X_t$, the goal is to train an image representation extractor that output a feature vector following the simulation feature distribution $x_t = f_t(im_t)\sim X_t = X_s$. In an adversarial fashion a discriminator D is introduced, it must distinguish if the input feature vector is extracted from $IM_s$ (outputs 1) or $IM_t$ (outputs 0).

To train $f_t$ to match the feature distribution $X_t$ to $X_s$, the following adversarial loss is used:  

\begin{equation}
\begin{tabular}{l}    
$L_{adv} = \mathbb{E}_{im_t \sim IM_t}[log(D(f_t(im_t)))].$
\end{tabular}
\label{eq:loss_fa1}
\end{equation}

On the other hand the discriminator is trained with the following loss:

\begin{eqnarray}
L_{dis} &=& -\mathbb{E}_{im_s \sim IM_s}[log(D(f_s(im_s)))] \\ &&-\mathbb{E}_{im_t \sim IM_t}[log(1 - D(f_t(im_t)))], \nonumber
\label{eq:loss_fa2}
\end{eqnarray}
where $f_s$ and $f_t$ share the same architecture and $f_t$ is initialized with $f_s$ weights. Training is performed by alternatively optimizing D and $f_t$. Every twenty epochs -- images are sampled from the simulator images dataset to balance the proportion of sim/real images being fed to the networks, a batch size of 50 is used.

\vspace{3mm}
Applying domain invariant feature learning implies that source and target domains feature spaces are identical so that the distributions can be aligned. Our office and the iGibson scene dataset are both indoor but do not share the same semantic space thus rendering the previous assumption partially violated. Furthermore the limited number of real images available restrain the scope of the transfer.

\subsection{Passage detector finetuning}
\vspace{4mm}

Prior decisions regarding the environment, the model and the generic nature of the task are supposed to facilitate any adaptation to the real world. Thus, transferring the passage detector simply consists in finetuning the model on the real life dataset. The six original labels are converted in binary labels, namely passage or non passage. Images are randomly split yielding 252 training images among which 134 are labelled as passage and 108 testing images among which 45 are labelled as passage. A binary cross entropy loss is used with an Adam optimizer with a 1e-4 learning rate. Training is performed on ten epochs with a batch size of five.

\vspace{3mm}
To gain more insight on the efficiency and necessity of the transfer multiple configurations are tested out with or without finetuning and with or without pretraining. As a baseline, result are computed for a Pytorch's Resnet18 backbone classifier pretrained on ImageNet1k and not finetuned on the real dataset (config \textbf{A}). The same pretrained network is evaluated after the finetuning task (config \textbf{B}). Finally the network previously trained on simulation to detect passage is evaluated with and without finetuning (resp. configs \textbf{C} and \textbf{D}). The following table \ref{tab:pd_t} shows the classification metrics on the test set.

\begin{table}[htbp]
\fontsize{10}{8}\selectfont
\centering
\caption{\bf Transfer passage detector performances on test set}
\begin{tabular}{ccccccc}
\hline
Configuration & acc & f1 & tn & fp & fn & tp \\
\hline
\textbf{A} & 0.49 & 0.07 & 51 & 12 & 43 & 2 \\
\textbf{B} & 0.8981 & 0.8842 & 55 & 8 & 3 & 42 \\
\textbf{C} & 0.6296 & 0.6875 & 24 & 39 & 1 & 44 \\
\textbf{D} & 0.8981 & 0.8866 & 54 & 9 & 2 & 43 \\
\hline
\end{tabular}
  \label{tab:pd_t}
\end{table}

As could be expected, without any finetuning the model pretrained for the passage detection task in simulation \textbf{C} significantly outperforms the model pretrained on ImageNet1k only \textbf{D}. However with only ten epochs of training the same model \textbf{C} yields equivalent results to the model pretrained on the passage detection task in simulation \textbf{D}. ImageNet1k comprises real images which could facilitate learning compared to the model trained on synthetic images. Training and testing images originate from the same scene, and given the low complexity of the dataset, ten epochs may be enough to recognize distinctive patterns about the scene only. Thus configuration \textbf{D} should still generalize better to any environment and is picked as the passage detector for the navigation trials.

\begin{figure}[htbp]
\centering
\includegraphics[width=0.7\linewidth]{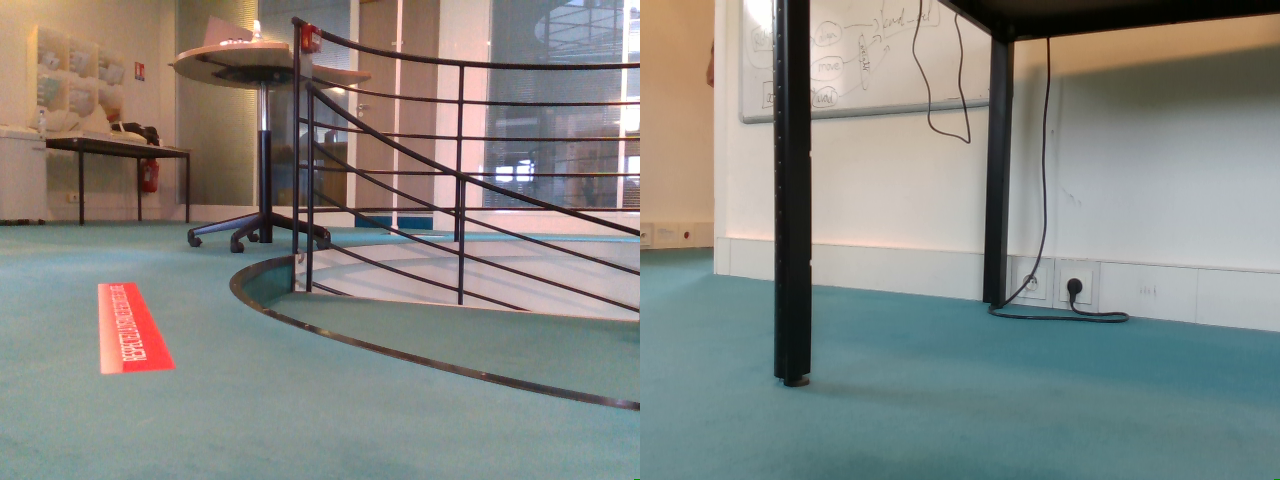}
\caption{False positives \textbf{C}}
\label{fig:pd_real_mistakes1}
\centering
\includegraphics[width=0.7\linewidth]{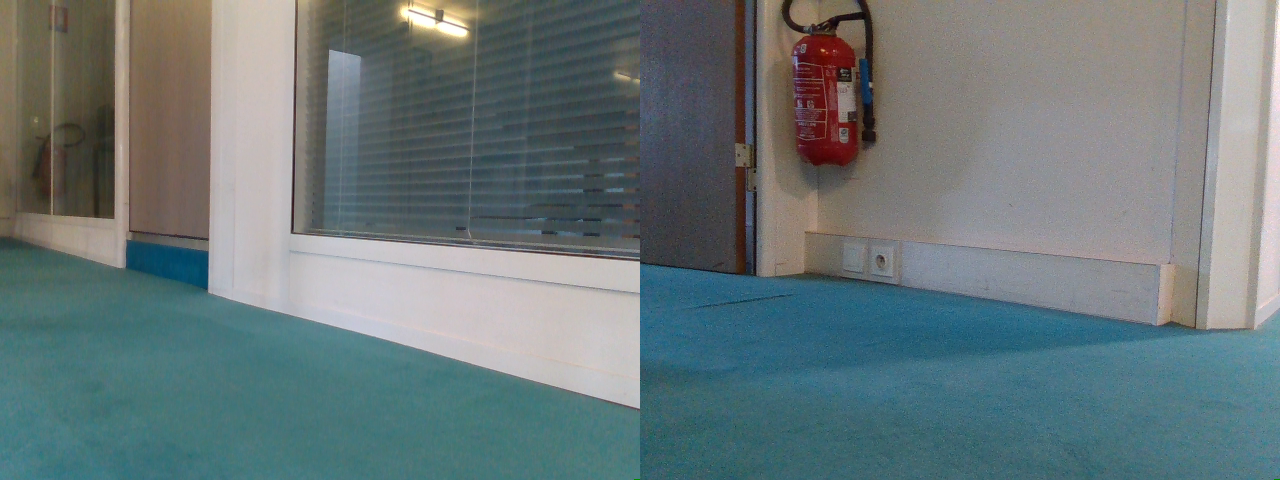}
\caption{False positives \textbf{D}}
\label{fig:pd_real_mistakes2}
\end{figure}

\vspace{3mm}
Overall, most of the errors on configurations \textbf{C} and \textbf{D} emanate from misguiding images without any clear label rather than fundamental mistakes ( as shown in  \ref{fig:pd_real_mistakes1},\ref{fig:pd_real_mistakes2}).

\subsection{Navigation}
\vspace{4mm}

\begin{figure}[htbp]
\centering
\includegraphics[width=1\linewidth]{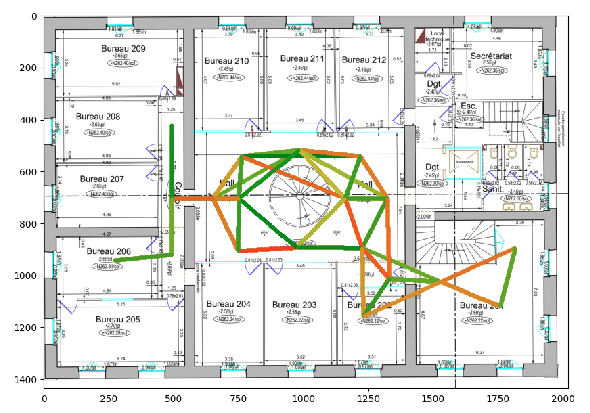}
\caption{Topological map}
\label{fig:real_topo_map}
\end{figure}

The last step toward real navigation consists in creating the topological map in a similar fashion as in a simulation. To that end the real dataset previously collected is once again used. Each position corresponds to a node in the graph. The connectivity matrix is manually designed by connecting adjacent neighbors, it is fed to the script in charge of creating the map. Then for each source/target edge, the angle in which to orientate while in the source position to reach the target position is computed ( in the absolute angle referential described in the dataset collection subsection). In the real dataset, the image with the corresponding source position number and closest absolute angle is processed through the finetuned passage detector to obtain the negative log-likelihood of passage associated to that edge. The resulting topological map is displayed in \ref{fig:real_topo_map}.

\vspace{3mm}
Given time and resources constraints the local policy was left as of, only upstream feature adaptation on the inputs was performed. From a software perspective very few adaptations are required on the core navigation script. Classes related to teleportation and episodes parser for the simulator are removed, some helper function dedicated to navigation monitoring are modified. Other than that, the only modification resides in loading the transferred navigation models instead of the synthetic ones. In practice, the navigation script is given the pretrained navigation models (local policy, image representation extractor, passage detector), the aforementioned topological map and the a real image of the goal and must reach the goal autonomously while avoiding collisions. To ensure that the goal is not mislocalized the goal image is directly sampled from the dataset.

\begin{figure}[htbp]
\centering
\includegraphics[width=.8\linewidth]{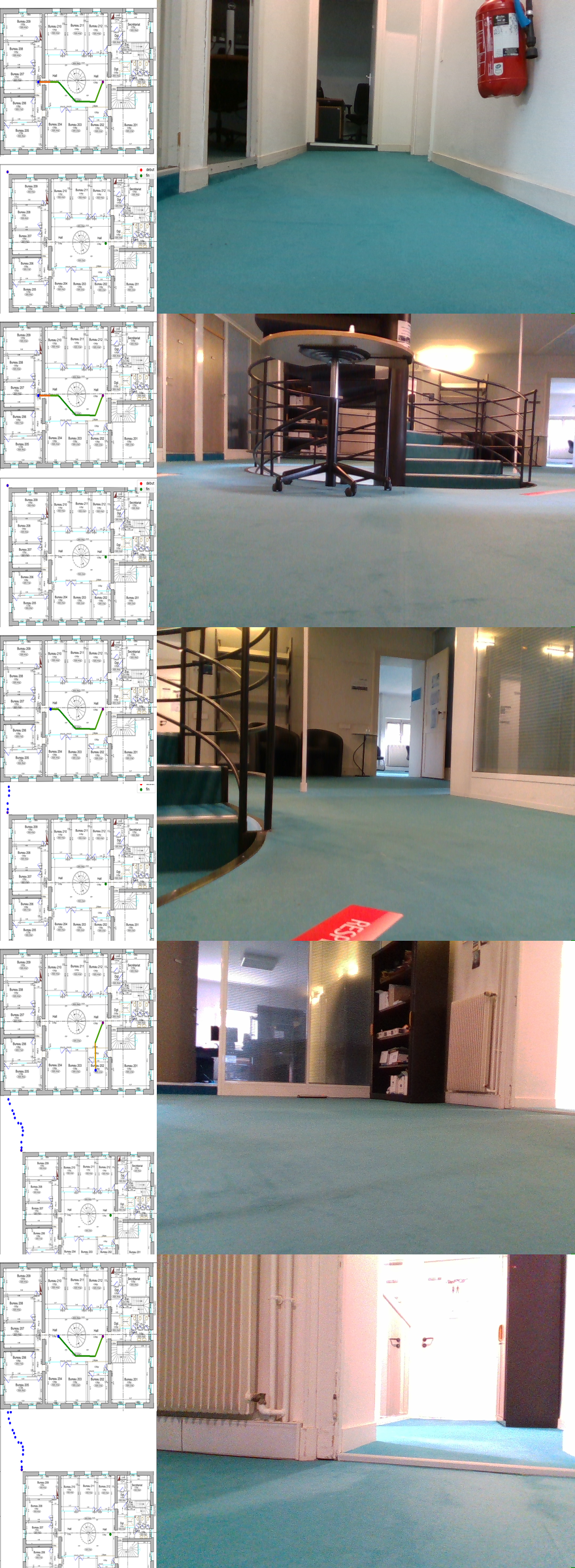}
\caption{Navigation episode (read top to bottom)}
\label{fig:realnav}
\end{figure}

\vspace{3mm}
Providing a quantitative evaluation of the navigation pipeline was too time consuming and could not be carried out throughout this internship. Therefore we provide a successful navigation example and some empirical remarks regarding the transfer task. In the example episode the agent is positioned in the western corridor and must reach the door leading to the elevator area on the other side of the central room. It implies crossing a door, avoiding the central staircase with curved trajectories. The euclidean distance between start and goal is 10.30 meters while the shortest path distance is around 13 meters. The goal was reached after 260 seconds of navigation without any collisions. \ref{fig:realnav} shows some consecutive images of the episode.

Localization was accurate most of the time except for some steps at the end of the episode. Those errors while not being critical increased the navigation time. Local policy showed a good behavior by capturing structural information about its surroundings. It accelerated when facing free space and decelerated when facing an obstacle or a wall. It also commanded relevant angular velocity commands to avoid imminent obstacles. The planning derived reliable trajectories allowing the agent to reach its destination within a reasonable time. Overall the errors that may have occurred were compensated by the other modules resulting in a successful episode.

\subsection{Observations}
\vspace{4mm}

As a conclusion this subsection lists the main issues faced after applying our transfer procedure. Other potential limitations such as local policy behaviors or topological map that are more related to the navigation pipeline rather than the transfer are not mentioned.

\begin{itemize}
    \item  Actuation comes as recurrent problem that weakens the navigation pipeline. The conversions between the numerical speed command and the actual command tension as motor input are completely different between the simulation agent and the real agent. Thus with the same speed command the real agent covers much more distance than its simulation counterpart. Furthermore, the non uniformity of the real carpet sometimes creates some slipping that combined with the bumpy nature of the floor disturbs the agent's movement. While being individually non that penalizing, those issues accumulate and contribute to the destabilization of the navigation. 
    \item Localization is also quite brittle and lacks robustness. It has shown extreme sensibility to change of luminosity between sunny and rainy days for example. In the semantically poor scene we operated in, it requires very distinctive features to be effective. Wide viewpoint change also destabilizes the localization more than it should.
    \item Due to technical limitations the local policy is not able to operate under optimal conditions. With relatively high velocity commands, some images end up being blurred. Combined with the fact that that the actual image topic refreshment rate is lower than the 10Hz the local policy was trained on, the input sequences do not carry as much information as it used to in simulation.    
    \item Thin objects remain hardly detectable under current operating conditions. Furthermore, low hanging flat object such as large furniture feet or cable cover, non present in simulation, are also hardly detectable due to their particular geometry. 
\end{itemize}

Despite these shortcomings we were able to carry out multiple successful navigation episodes on the scene. It demonstrates that the choice of the framework and solutions to tailor and train our navigation pipeline did allow an easy transfer with a limited amount of work to be made.

\section{Conclusion}
This internship aimed at creating, in a simulation, a navigation pipeline whose transfer to the real world could be done with as few efforts as possible. The emphasis was rather put on studying the sim2real gap which is one the major bottlenecks of modern robotics and autonomous navigation. The internship was divided into two consecutive parts, designing and training this navigation pipeline followed by tackling the transfer with the aim of making our physical agent autonomously navigate in a purposefully selected environment. 

All the upstream decisions regarding the choice of the environment, the navigation framework and the design of the navigation pipeline aimed at facilitating the subsequent transfer. A topological approach to tackle space representation was picked over metric approaches because they generalize better to new environments and are less sensitive to change of conditions. The navigation pipeline was decomposed as a localization module, a planning module and a local navigation module operating under iGibson through ROS. These modules utilize three different networks, a image representation extractor, a passage detector and a local policy. The latter are trained on specifically tailored tasks with some associated dataset created for those specific tasks. Localization is tackled as an image retrieval task using a deep neural network trained on an auxiliary task as a feature descriptor extractor. The local policy is trained with behavioral cloning from expert trajectories gathered with ROS navigation stack. When evaluated over a subset of scene on a image goal navigation, the navigation pipeline yields satisfactory performances although the collision rate increases significantly when the scene becomes too cluttered. Performances could be improved by replacing behavioral cloning by a more complex imitation learning approach to train the local policy. Integrating or adding some finer solutions to avoid furniture would certainly be beneficial. Finally, using a single topological map presents structural limitations that could be alleviated if the latter was used in a hierarchical way or combined with metric representation.   

The second floor of NLE's castle was picked as the real navigation scene. As an office, the scene is semantically poor with high similarities between rooms. The transfer procedure first consisted in collecting enough images. The modularity of the pipeline permitted individual transfer of the different models. The passage detector was simply finetuned on a few hundred real images yielding very good accuracy on the scene. Some feature adaptation was applied to learn a real image representation extractor whose output feature distribution is the same as in simulation. Local policy was left unmodified and very few changes had to be made on the script and environments. After creating a topological map of the scene from the collected dataset, the TTbot agent was able to navigate autonomously within the scene. We were not able to provide a quantitative and thorough evaluation, however from experience some conclusions can be derived. The local policy while displaying a similar behavior as in simulation did struggle avoiding small obstacles non necessarily present in simulation, having no common optimal policy between simulation and reality is a recurrent problematic. Multiple actuation issues destabilized the navigation, the lack of a global framework to address actuation transfer issues make it difficult to tackle this problem. Localization while effective in simulation proved very sensitive to viewpoint changes and illumination conditions calling for a much more sophisticated approach. Despite those limitations the agent was able to successfully complete a significant number of episodes from this basic transfer procedure. The aforementioned issues would still have to be addressed to significantly improve reliability and efficiency.  

\clearpage
\bibliographystyle{unsrt}
\bibliography{biblio}

\end{document}